\pdfoutput=1

\documentclass[11pt]{article}

\usepackage[]{emnlp2023}

\usepackage{times}
\usepackage{latexsym}

\usepackage[T1]{fontenc}

\usepackage[utf8]{inputenc}

\usepackage{microtype}

\usepackage{inconsolata}

\usepackage{comment}
\usepackage{ifthen} 
\usepackage{overpic}
\usepackage{booktabs}
\usepackage{multirow}
\usepackage{rotating}
\usepackage{wrapfig}
\usepackage{amsmath}
\usepackage{amssymb}
\usepackage{bbm}
\usepackage{dsfont}
\usepackage{wrapfig}
\usepackage{verbatim} 
\usepackage{nicefrac} 
\usepackage{caption}
\usepackage{pifont}
\usepackage{colortbl} 
\usepackage{color}
\usepackage{xcolor}
\usepackage{balance}
\usepackage{marvosym}

\usepackage{float}

\newcommand{\tabincell}[2]{\begin{tabular}{@{}#1@{}}#2\end{tabular}}

\newcommand{\mathmat}[1]{\mathbf{#1}}

\newcommand{\secref}[1]{Section \ref{#1}}
\newcommand{\tabref}[1]{Table \ref{#1}}
\newcommand{\figref}[1]{Figure \ref{#1}}

\newcommand{\appref}[1]{Appendix \ref{#1}}

\def\ie{\emph{i.e.}}
\def\eg{\emph{e.g.}}
\def\etc{\emph{etc}}

\title{Shikra: Unleashing Multimodal LLM's Referential Dialogue Magic}

\author{Keqin Chen$^{12}$\thanks{~~Work done during internship at SenseTime Research}, Zhao Zhang$^{1}$\thanks{~~Equal Contribution \& Project Leader}, Weili Zeng$^3$, Richong Zhang$^2$, Feng Zhu$^1$, Rui Zhao$^{14}$ \\
 $^1$SenseTime Research; $^2$SKLSDE, Beihang University \\\{$^3$SEIEE, $^4$Qing Yuan Research Institute\}, Shanghai Jiao Tong University \\
\texttt{chenkq@act.buaa.edu.cn}; \texttt{zzhang@mail.nankai.edu.cn}
}

\begin{document}
\maketitle

\begin{abstract}
In human conversations, individuals can indicate relevant regions within a scene while addressing others. In turn, the other person can then respond by referring to specific regions if necessary. This natural referential ability in dialogue remains absent in current Multimodal Large Language Models (MLLMs).
To fill this gap, this paper proposes an MLLM called Shikra, which can handle spatial coordinate inputs and outputs in natural language. 
Its architecture consists of a vision encoder, an alignment layer, and a LLM. It is designed to be straightforward and simple, \textbf{\textit{without}} the need for extra vocabularies, position encoder, pre-/post-detection modules, or external plug-in models.
All inputs and outputs are in natural language form.
Referential dialogue is a superset of various vision-language (VL) tasks. Shikra can naturally handle location-related tasks like REC and PointQA, as well as conventional VL tasks such as Image Captioning and VQA. Experimental results showcase Shikra's promising performance.
Furthermore, it enables numerous exciting applications, like providing mentioned objects' coordinates in chains of thoughts and comparing user-pointed regions similarities. Our code and model are accessed at \url{https://github.com/shikras/shikra}.
\end{abstract}

\section{Introduction}
\label{sec:intro}
\begin{figure}[t]
	\begin{center}
		\includegraphics[width=1.0\linewidth]{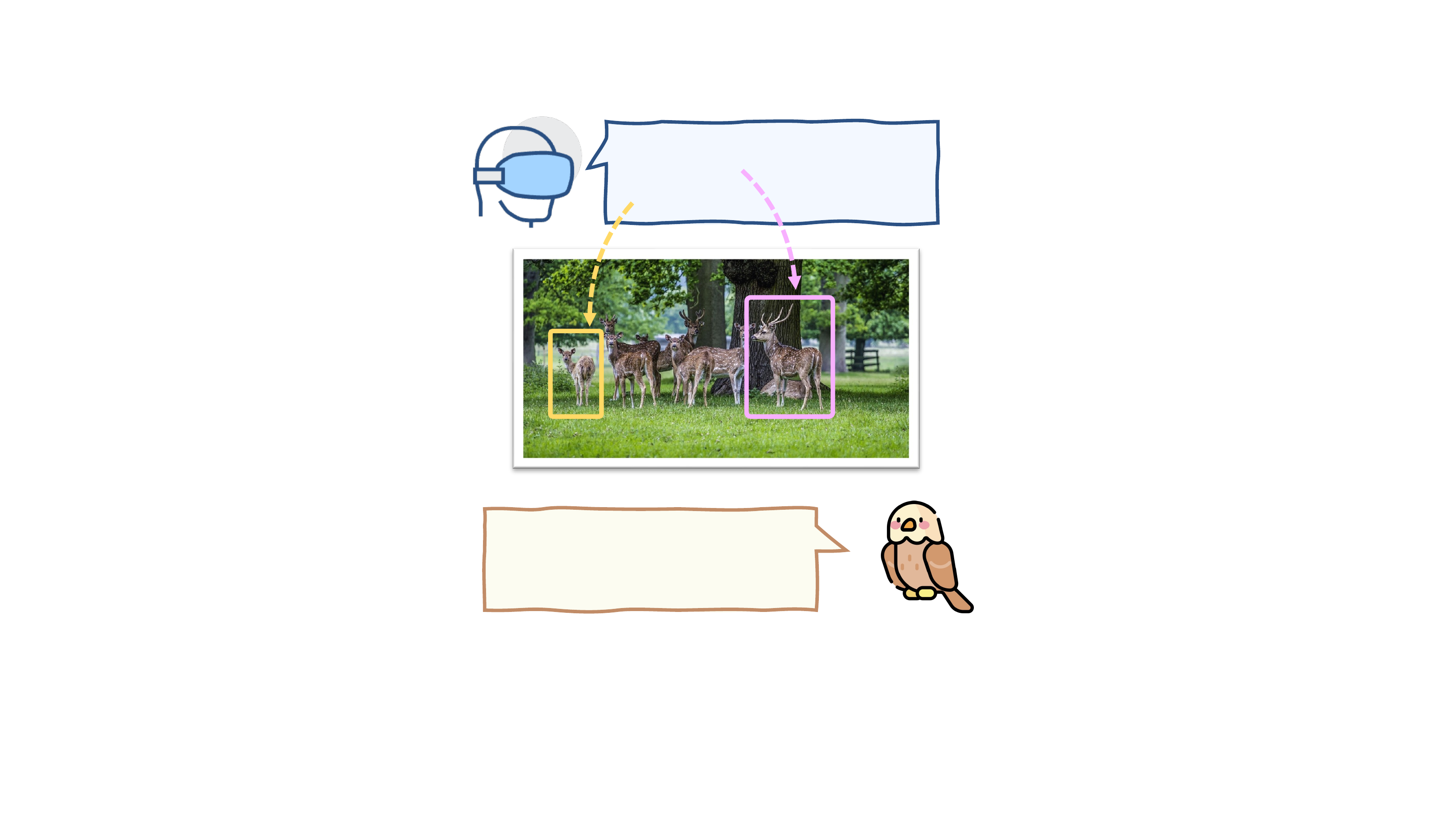}
	\end{center}
    \caption{
    \textbf{Demo of Referential Dialogue (RD)}. Users can point to specific areas and ask questions. In turn, Shikra will indicate the specific regions when replying, if necessary. More interesting dialogues can be found in \figref{fig:cases} and \appref{sec:more_cases}.
    } 
    \vspace{-8pt}
\label{fig:teaser}
\end{figure}
In recent months, Multimodal Large Language Models (MLLMs) have witness remarkable progress \citep{alayrac2022flamingo,kosmos,liu2023llava,zhu2023minigpt,li2023otter,gao2023la_v2,dai2023instructblip}. They brings eyes to Large Language Models (LLMs), where users can talk about the input image.
However, although these models can perceive image content, they cannot engage in dialogue with users regarding the precise positions of the content. Users cannot indicate areas of interest in the image, and the models cannot provide the exact locations of the described content.
Differently, as shown in \figref{fig:teaser}, in human daily communication, different regions or objects in the scene are often attended to, and people can speak and point to these regions for efficient information exchange.
We refer to this interaction mode as Referential Dialogue (RD).
If an MLLM excels in this skill, it will bring numerous exciting applications. 
For instance, applying it to Mixed Reality (XR) headsets like Apple Vision Pro, users can indicate anything to converse with the AI assistant. The AI assistant can display the prompt area in the field of view when necessary.
It also assists visual robots in communicating with individuals by comprehending their specific reference positions. It facilitates online shopping by enabling users to inquire about items of interest in an image.

\begin{figure*}[!t]
    \begin{center}
        \includegraphics[width=0.97\textwidth]{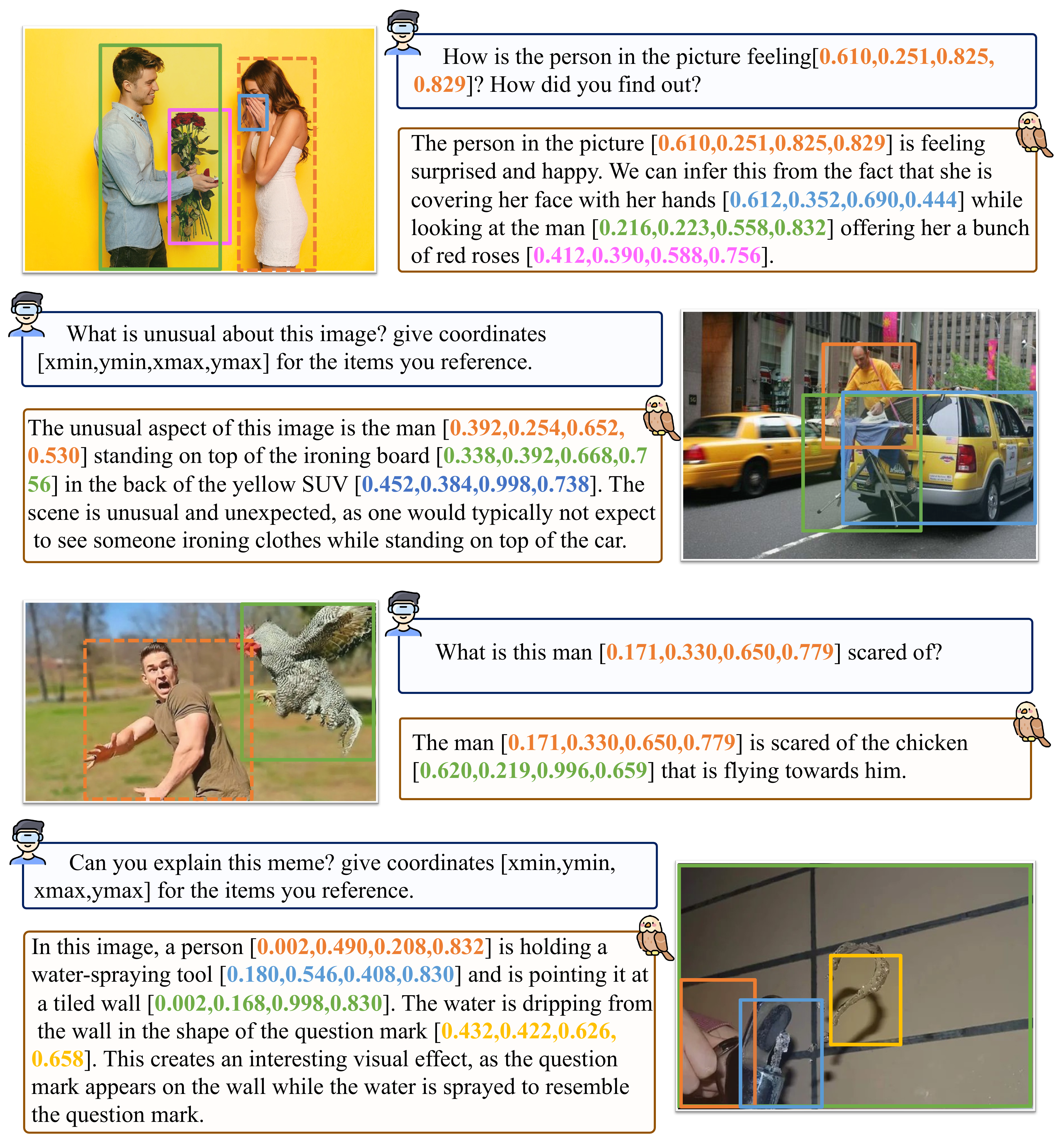}
    \end{center}
   \caption{\textbf{Referential Dialogues between real users and Shikra-7B.} The dashed box on an image represents the area referred to by the user or jointly referred to by Shikra, while the solid box represents the area solely referred to by Shikra. More RD results and applications on conventional VL tasks can be found in \appref{sec:more_cases}.}
\label{fig:cases}
\end{figure*}

In this paper, we evolve MLLM to open the veil of referential dialogue. We create Shikra\footnote{Shikra is a hunter's companion, capable of understanding human language and gesture instructions, and locating and capturing prey in the wild.}, a unified model capable of handling inputs and outputs of spatial coordinates. All coordinates, both input and output, are represented in natural language numerical form without introducing any extra vocabularies or position encoder. The Shikra architecture comprises a vision encoder, an alignment layer, and a LLM. We do \textbf{not} introduce any pre-/post-detection modules or external plug-in models, making Shikra unified and simple.
We provide several real conversations with users in the \figref{fig:cases} and \appref{sec:more_cases},
where users can use it to compare the differences of multiple regions, inquire about the meaning of the thumbnail, discuss specific objects, \etc.
Shikra can provide explanations when answering any question, not only verbally but also spatially.

Referential dialogue is a superset of many vision-language (VL) tasks. Shikra, skilled in RD, can naturally work on these tasks with promising performance, including Visual Question Answering (VQA), image captioning, and location-related tasks such as Referring Expression Comprehension (REC) and PointQA,  We illustrate some of them in \figref{fig:cases}. For more quantitative results, please refer to \secref{sec:experiment_vltask}.
Besides, this paper also addresses intriguing questions, such as how to represent position in an image (\secref{sec:loc}). Do previous MLLMs possess the capability to comprehend absolute positions? (\secref{sec:chessboard}). Can the reasoning process with location information assist in providing more accurate answers to questions? (\secref{sec:clevr}).
We hope that these analysis experiment can inspire future research on MLLMs.

The main contributions of this paper are:
\begin{itemize}
    \item This paper introduces the task of Referential Dialogue (RD), which is an essential component of everyday human communication and possesses extensive practical applications.
    \item We present Shikra, a generalist MLLM, for RD.
    Shikra is simple and unified, \textbf{without} introducing extra vocabularies, pre-/post-detection module, or external plug-in models. 
    \item  Shikra handles unseen settings effortlessly, creating diverse application scenarios. It also achieves promising performance on conventional visual language tasks such as REC, PointQA, VQA, and Image Captioning, without finetuning.
\end{itemize}

\section{Related Works}
\label{sec:related_work}
\subsection{Multimodal Large Language Model}
Expanding the large language model to a multimodal version has garnered widespread attention. Flamingo \citep{alayrac2022flamingo} integrates visual adaption layers (like Perceiver) to an LLM, and trained on a large-scaled interleaved image-text dataset. OpenFlamingo \citep{anas2023OpenFlamingo}  re-implements Flamingo and releases it to the community along with an M3C dataset. Subsequently, MM-GPT \citep{gong2023mmgpt},
and Otter \citep{li2023otter} tune on carefully constructed instruction data for a more user-friendly interaction.
Another genre is BLIP-2 \citep{li2023blip2},
which align queried visual feature with text using multiple vision-language losses (model named Q-Former), and tunes a simple fully connection layer to feed the queried embedding to a frozen language model.
Mini-GPT4 \citep{zhu2023minigpt}, mPLUG-OWL \citep{ye2023mplug}, VPGTrans \citep{zhang2023transfer}, and InstructBLIP \citep{dai2023instructblip} retain Q-Former, replace language model to a larger one, and then tuning on meticulously collected instruction data.
Additionally, there are simpler and more direct methods:
FROMAGe \citep{koh2023fromge} and LLaVA \citep{liu2023llava} directly feed visual features to the LLM using only a learnable fully connected layer.
The closed source business model GPT-4 \citep{openai2023gpt4} also demonstrates astonishing image comprehension capabilities.
Recently, interesting works have made remarkable progress by extending LLM to audio, \eg, KOSMOS-1 \citep{kosmos}, X-LLM \citep{chen2023xllm}, PandaGPT \citep{su2023pandagpt} and control systems like
PaLM-E \citep{driess2023palme} and EmbodiedGPT \citep{mu2023embodiedgpt}

\subsection{Vision-Language Positioning Tasks}
Many vision-language tasks require localization representation. 
\textbf{Tasks with output boxes}: 
Referring Expression Comprehension (REC) \citep{kazemzadeh2014refcoco,mao2016refcocog} aims to localize a target object in an image described by a referring expression.

Described Object Detection \citep{xie2023dod} extends REC to more realistic scenarios where the object may not exist or there may be multiple objects.
VQA Grounding aims to answer visual questions and associate the answers with specific visual regions or objects.
\textbf{Tasks with input boxes}: 
Given an image and a location box, the task of Grounding Caption (GC) \citep{zhou2020more} is to generate a description for this location by considering the surrounding environment.
Compared to GC, Referring Expression Generation (REG) \citep{liu2017referring} requires the generated description to indicate that it describes this region specifically, not others, making it necessary for the description to be discriminative.
PointQA \cite{mani2020pointqa} requires a model answer for a visual question where the questioner queries a specific position in the picture.
\textbf{Differently}, our model is not only compatible with the above tasks, but also can handles the input and output of position representation flexibly and simultaneously, bringing Referential Dialogue and extending new dimensions to positional tasks.

\subsection{Position Representation}
\label{sec:pos_rep}
\textbf{Inputting} regions of interest into the model presents various approaches.
Some methods \citep{bracha2023disclip} directly concatenate cropped image patches with the original image as model input.
There are also some methods \citep{lin2020fca,lin2022mmiis} that use 0/1 mask or Gaussian map input with the original image to emphasize the area of user interest.
Some methods \citep{tancik2020fourier,kirillov2023sam} first encode points and boxes to positional encodings then add them to intermediate features or learned queries.
\textbf{Outputting} regions of interest is a highly focused technique,
existing many positioning paradigms . 
Anchor-based methods utilize predefined sliding windows and proposal candidate regions for classification., \eg, Fast R-CNN \citep{girshick2015fastrcnn}.
Some one-stage methods remove anchors and directly regress four values for bounding box coordinates, \eg, FCOS \citep{tian2019fcos}.
Some methods adopt one-to-one label assignment to evolve object detection into an end-to-end manner, \eg,  DETR \citep{carion2020detr} and POTP \citep{wang2021end}.
An interesting genre is Pix2seq \citep{chen2021pix2seq},
which formalizes the detection task as a sequence generation task. It desires the spatial position of the image in 1,000 bins and uses a 1,000-token vocabulary to represent it.
For detection, Pix2seq performs classification on the coordinate vocabulary in an auto-regressive manner.
Following Pix2seq, several methods, \eg, OFA \citep{wang2022ofa},
Unified-IO \citep{lu2022unified},
UniTab \citep{yang2022unitab},
GIT \citep{wang2022git}, and VisionLLM \citep{wang2023visionllm} introduce similar coordinate vocabulary alongside the language vocabulary for object detection and REC tasks.
\textbf{Differently}, Shikra formulates position input/output as the most natural and flexible form of language and compare it with the extra coordinate vocabulary in \secref{sec:loc}.

\section{Referential Dialogue} 
\label{sec:dialogue}
To better understand the interesting abilities of our model, we demonstrated real users' communications in  \figref{fig:teaser} and \figref{fig:cases}.
As shown in the first demo of \figref{fig:teaser}, the user points to two deer, and inquires, ``\textit{What is the difference between this deer and another deer?}'' 
When Shikra answered, she not only mention the differences but also output the coordinates of the differences.
The subsequent examples in \figref{fig:cases} are alike.
To our knowledge, there have been no unified models that can achieve such functionality before.
RD is a superset of numerous vision-language tasks. Shikra can perform most tasks like current MLLM, including VQA, Image Caption, and multimodal dialogue. Furthermore, it handles tasks that they cannot, like REC, REG, and PointQA. The model demonstrates proficiency in tasks not in the training set, such as identifying similarities between two indicated objects, or counting somethings, and providing their positions.
We show more results in \appref{sec:more_cases}.
If you are interested in quantitative experiments, you can refer to \secref{sec:experiment} later.

\section{Chessboard Test for Current MLLM} 
\label{sec:chessboard}
Can the current MLLM model understand absolute spatial positions?
The current MLLMs cannot directly output coordinates; thus,
in this section, we designed a chessboard test, which simplifies the object grounding into a part choice task.
Specifically, we divide a image into a $2\times2$ chessboard.
Next, we ask, ``\textit{<image> Which part is <expr> in if the picture is divided equally into four 2 by 2 parts? Choose from: (A) Top-left (B) Top-right (C) Bottom-left (D) Bottom-right.}'',
where <image> and <expr> denote input image tokens and Class name.
We construct test data from LVIS \cite{gupta2019lvis}, which is a perception detection with over 1000 entry-level object categories. We choose objects that are completely within a certain part (\ie, ambiguous positions are not considered).
In total, we select 600 images per part, resulting in 2,400 images across 945 categories.
We employ LLaVA-13B \citep{liu2023llava} for the chessboard test
, but the results are unsatisfactory. We tried various instruction methods, and LLaVA should achieve an accuracy of 25.96\%, which is comparable to random selection.
This suggests that prior coarse-grained vision-language alignment pre-training may be inadequate for MLLMs to capture the exact spatial position of an image. We need to explore appropriate coordinate representations and finer-grained training data.

\section{Breeding Shikra}
\label{sec:tuning}
This section introduces the birth of Shikra, encompassing its structure design, position representation, training data construction, and training strategies.
\subsection{Architecture}
We selected the pre-trained ViT-L/14 of CLIP as visual encoder and Vicuna-7/13B as our LLM. We use one fully connected layer to map the ViT's $16\times16\times$ output embedding $\mathmat{V}\in\mathbb{R}^{16\times 16\times 1024}$ to $\mathmat{V'}\in\mathbb{R}^{256\times D}$ for modal alignment and correct input dimension of LLM. $D$  is 4,096 for Vicuna-7B and 5,120 for Vicuna-13B.
Visual embedding can be inserted into anywhere of input sequence.
During training, both the fully connected layer and the entire language model are involved. 
We do not introduce any vocabulary or special encoder for encoding position information. We have not introduced additional pre-/post-detectors for points or bounding boxes.
The model using Vicuna-7B is called Shikra-7B, and the other, using Vicuna-13B, is named Shikra-13B.


\subsection{Numerical representation of position}
We represent the position using numerical values in Natural Language in a highly intuitive manner.
We use $[x_\text{min},y_\text{min},x_\text{max},y_\text{max}]$ to denote the bounding box and $[x_\text{center},y_\text{center}]$ to denote region center point.
$x$ and $y$ is normalized according to the size of the image. We default to keeping 3 decimal places for each number.
These coordinates can appear anywhere in the input and output sequence of the model.
For example, User Question: ``\textit{How many other clothes in the <image> are of the same color as the jacket $[0.268, 0.372]$?}''. Shikra reply: ``\textit{The jacket $[0.268, 0.372]$ is green. We can find a T-shirt $[0.653, 0.532]$ and cropped pants $[0.569, 0.101]$ a with same green color. So the answer is two.}''
The square brackets that record coordinates naturally appear in sentences and can serve as any sentence component.
Like regular text, tokenizing without discrimination.

\subsection{Instruction data construction}
We utilize two types of data to train Shikra: the reorganized public datasets, and the high-quality RD data built from Flickr30K Entities \citep{plummer2015flickr30ke} using GPT-4 \citep{openai2023gpt4}.

\subsubsection{Reorganization of public data}
\label{sec:reorg}
We collection training data from public VQA, Image Captioning datset, and several datasets already containing positional annotation, such as RefCOCO \citep{kazemzadeh2014refcoco} for REC/REG, visual gemone \citep{krishna2017visualgenome} for grounding caption, Visual-7W \citep{mani2020pointqa} for PointQA.
We also define new task forms, such as Spotting Captioning, which requires the model to describe the image and spots the mentioned objects or regions using points or boxes. We use Flickr30K Entities for this task. All the data used and corresponding tasks can be found in \appref{sec:train_data}. Note that all the data used were included in the reported model results, unless stated otherwise for specific comparative experiments.  Additionally, it should be mentioned that we have \textbf{excluded} images present in the test and validation data from the training data to prevent potential data leakage, despite their distinction in terms of image-text pairs. 

\subsubsection{Generated data}
\label{sec:gendata}
The existing publicly available data is not sufficient to train an MLLM skilled in RD, as they lack CoT data with positional annotations, natural communication data with positional annotations, \etc. We resort to GPT-4 to obtain high-quality RD annotations from Flickr30K Entities. Flickr30K Entities has five descriptions for each image. These mentioned objects appearing in the image will be labeled using bounding box. Although the API of GPT-4 temporarily \textbf{cannot} see images, we explained the format of the bounding boxes to GPT-4 and asked it to understand the image through these five sentences and boxes. Next, we require GPT-4 to design Q\&A pairs. When designing problems, these questions must be able to determine answers from known information. In this way, we generated 5,922 QA pairs, where coordinate information may appear in both questions and answers. The dataset will continue expanding in the future.  You can refer to it as Shikra-RD.

\subsubsection{Task prompts}
We construct variable task templates for different tasks. For instance, for the spottingS caption task, we can use ``\textit{Can you provide a description of the image <image> and include the coordinates} [x0,y0,x1,y1] \textit{for each mentioned object?}'' where <image> represents the visual tokens. For PointQA, we can use ``\textit{Referring to point <objs> in image <image>, give a direct answer to '<question>'}'' where <objs> denotes the coordinates of the region and <question> represents the question from the source dataset. For REC, ``\textit{In <image>, I need the bounding box coordinates of <expr>.}'' where <expr> is the expression. More templates for different tasks can be found in the Appendix.

It should be noted that we cannot use an invariant task template for a specific type of task. In this case, the model cannot flexibly accept user instructions. To solve this problem, we first describe the purpose of specific tasks, write a sample template, and then have GPT-4 rewrite it in rich language, expanding it into hundreds of variations to convey the same meaning. During training, we can randomly choose from them. 
We provide details on some generated task templates in the \appref{sec:task_temp}.

\subsection{Tuning details}
Shikra is trained in two stages. In the first stage, we train it on the reorganized VL dataset (\secref{sec:reorg}) for 100,000 steps (around 1.5 epoch); In the second stage, we
raise the sampling ratio to 50\% on LLaVA-Instruct-150K \citep{liu2023llava} and our generated RD data (\secref{sec:gendata}).
In both stages, we freeze the visual encoder and tune all parameters in LLM. We adopt AdamW \citep{DBLP:conf/iclr/LoshchilovH19adamw} as the optimizer and cosine annealing scheduler \citep{DBLP:conf/iclr/LoshchilovH17cos} as learning rate scheduler with an initial learning rate of 2e-5 and global batch size of 64. All training runs on 8 NVIDIA A100 GPUs. It takes around 100h for stage one training and 20h for stage two.

\section{Experiment and Analysis}
\label{sec:experiment}
\begin{table}[t]
\centering
\renewcommand{\tabcolsep}{3mm}
\caption{\textbf{Comparing different forms of CoTs.} We train three toy models of Shikra-7B (without using additional datasets) on the CLEVR dataset. Q, A, C, and $\text{C}^\text{Point}$ denote the \textbf{Q}uestion, final \textbf{A}nswer, \textbf{C}hain of thoughts, and \textbf{C}hain of thoughts with \textbf{P}ointing.}
\begin{tabular}{ccc}
\toprule
Q$\rightarrow$A & Q$\rightarrow$CA & Q$\rightarrow$$\text{C}^\text{Point}$A  \\
\cmidrule(lr){1-3}
88.07 & 80.68 & 93.97   \\
\bottomrule
\end{tabular}%
\label{tab:clevr}
\end{table}
\begin{table}[t]
\centering
\caption{\textbf{Comparing different position representations.}
We implement Shikra-7B in two different representation forms and train two toy models solely on RefCOCO, RefCOCO+/g, and Visual Genome for controllable comparison. 
Vocab. means to use extra vocabularies to represent coordinates, like \citep{chen2021pix2seq,wang2022ofa}, and Numerical means to directly use numerals in natural language to express coordinates.} 
\begin{tabular}{l|l|cc}
\toprule
Dataset                  & Split & Vocab. & Numerical \\ 
\cmidrule(lr){1-4}
\multirow{3}{*}{RefCOCO}  &   val     & 81.03 & 81.47 \\
                          &  test-A   & 86.94 & 87.40 \\
                          &  test-B   & 70.91 & 73.25 \\ \cmidrule(lr){1-4}
\multirow{3}{*}{RefCOCO+} &   val     & 72.32 & 74.30 \\
                          &  test-A   & 81.78 & 83.29  \\
                          &  test-B   & 59.95 & 63.08 \\ \cmidrule(lr){1-4}
\multirow{2}{*}{RefCOCOg} &  val-u    & 72.81 & 75.69 \\
                          &  test-u   & 73.78 & 75.52 \\
\bottomrule
\end{tabular}
\label{tab:numer}
\end{table}
\subsection{Grounding CoT or verbal CoT?}
\label{sec:clevr}
The process of providing reasoning before giving an answer is called Chain of the thoughts (CoT), which provides good explanatory during model judgments. However, CoT often suffer from hallucinations \citep{zhang2023mmcot}, which often do not improve the performance of the final answer.
Current MLLMs are also suffer from serious visual hallucination \citep{li2023obj_Hallucination}.
In this section, we investigate whether CoT with position annotations can reduce hallucinations and improve model performance. In this paper, we refer to this type of CoT as Grounding CoT (GCoT).  We train our Shikra-7B (without pre-training) on CLEVR~\citep{johnson2017clevr} in three settings: 1) Only use Question and Answer (Q$\rightarrow$A); 2) Use Question, CoT, and answer (Q$\rightarrow$CA); 3) Use GCoT with Center Point annotation and answer (Q$\rightarrow$$\text{C}^\text{Point}$A). We record they performance in \tabref{tab:clevr}.
Using only CoT to train the model (Q$\rightarrow$CA) and requiring a reasoning process before the final answer decreases performance compared to direct answering setting (Q$\rightarrow$A). 
In the Q$\rightarrow$$\text{C}^\text{Point}$A setting, we ask the model to provide CoT along with center points $[x_\text{center},y_\text{center}]$ for each mentioned object. Performance improved by 13 points compared to Q$\rightarrow$CA and 5.9 points compared to Q$\rightarrow$A, indicating that training with positional annotations suppresses visual hallucination.
This is a preliminary attempt at GCoT, and it is a promising direction worth exploring.

\subsection{Location tokens or just numbers?}
\label{sec:loc}

\begin{table*}[t]
\centering
\caption{\textbf{Results on standard REC task}. Generalist VL model
Generalist VL models can directly perform various vision-language tasks, including image captioning, VQA, REC, \etc.
Specialist models are those specifically designed for localization tasks (\eg, UNINEXT, \citealp{yan2023uninext} and G-DINO, \citealp{liu2023gdino}), or generalist pretraining models that have undergone multitask localization finetuning (\eg, \citealp{yang2022unitab}) or single-task finetuning (\eg, \citealp{wang2022ofa}).
We select the three current best performing models \cite{liu2023gdino,yan2023uninext,wang2023one_peace} as baselines.
OFA-L* \citep{wang2022ofa} refers to the OFA-Large checkpoint without finetuning.
GRIT refexp is the ablation split \citep{lu2022unified}.
}
\label{tab:rec}
\resizebox{\textwidth}{!}{%
\begin{tabular}{l|l|ccccccccc}
\toprule
\multirow{2}{*}{Model type}
& \multirow{2}{*}{Model}  & \multicolumn{3}{c}{RefCOCO} & \multicolumn{3}{c}{RefCOCO+} & \multicolumn{2}{c}{RefCOCOg} & GRIT \\
 &  & val & test-A & test-B & val & test-A & test-B & val-u & test-u & refexp \\
\cmidrule(lr){1-2}\cmidrule(lr){3-11}
\multirow{5}{*}{\tabincell{l}{Generalist VL SOTAs \\ (w/o finetuning)}}
& GPV-2       & - & - & - & - & - & - & - & - & 51.50\\
& OFA-L*      & 79.96 & 83.67 & 76.39 & 68.29 & 76.00 & 61.75 & 67.57 & 67.58 & 61.70 \\
& Unified-IO  & - & - & - & - & - & - & - & - & 78.60\\
& OFASys      & - & 80.10 & - & - & - & - & - & - & -\\
& VisionLLM-H   & - & 86.70     & - & - & - & - & - & - & -\\
& \textbf{Shikra-7B}      & 87.01 & 90.61 & 80.24 & 81.60 & 87.36 & 72.12 & 82.27 & 82.19 & 69.34\\
& \textbf{Shikra-13B}      & 87.83 & 91.11 & 81.81 & 82.89 & 87.79 & 74.41 & 82.64 & 83.16 & 69.03\\
\cmidrule(lr){1-2}\cmidrule(lr){3-11}
\multirow{3}{*}{\tabincell{l}{Specialist SOTAs \\ (Specialist/Finetuned)}}
& G-DINO-L     & 90.56 & 93.19 & 88.24 & 82.75 & 88.95 & 75.92 & 86.13 & 87.02 & -\\
& UNINEXT-H    & 92.64 & 94.33 & 91.46 & 85.24 & 89.63 & 79.79 & 88.73 & 89.37 & -\\
& ONE-PEACE    & 92.58 & 94.18 & 89.26 & 88.77 & 92.21 & 83.23 & 89.22 & 89.27 & -\\
\bottomrule
\end{tabular}%
}
\end{table*}
\begin{table}[t]
\centering
\renewcommand{\tabcolsep}{1.2mm}
\caption{\textbf{Comparing pointQA capabilities on the Visual-7W} \citep{zhu2016v7w}. Visual-7W features a `which box' setting, requiring the model to select one matching box from four options based on the given description.  Accuracy (\%) is used for evaluation.
}
\begin{tabular}{cccc|c}
\toprule
\citeauthor{zhu2016v7w} & \citeauthor{hu2017v7w_method1} & \citeauthor{lu202012in1} & \citeauthor{lu202012in1}* & Shikra \\
\cmidrule(lr){1-5}
56.10 & 72.53 & 82.75 & 83.35 & 85.33  \\
\bottomrule
\end{tabular}%
\label{tab:experiment_v7w}
\end{table}
\begin{table}[t]
\centering
\renewcommand{\arraystretch}{1.23}
\renewcommand{\tabcolsep}{0.845mm}
\caption{\textbf{Comparing pointQA capabilities  on the LookTwice-QA} \citep{mani2020pointqa}, where the models are asked to answer question based on the input point/box. Pronoun, Superclass (Super cls.), and Class indicate different levels of referential clarity in the question, \eg, ``How many of these [$\varnothing$/fruits/apples] <obj>?" We use Shikra-13B and Accuracy (\%) for evaluation.
}
\begin{tabular}{l|cc|cc}
\toprule
Type & \multicolumn{2}{c|}{Point} & \multicolumn{2}{c}{Box} \\
Model & \citeauthor{mani2020pointqa} & Shikra & \citeauthor{mani2020pointqa} & Shikra \\ \cmidrule(lr){1-5}
Pronoun     & 56.5 & 70.0 & 60.2 & 70.3  \\
Super cls.  & 59.1 & 70.2 & 59.8 & 71.4  \\
Class       & 62.8 & 71.8 & 61.4 & 72.3  \\
\bottomrule
\end{tabular}%
\label{tab:twiceqa}
\end{table}
\begin{table*}[h]
\centering
\renewcommand{\tabcolsep}{0.9mm}
\caption{\textbf{Comparing generalist models on VQA and Image Captioning.}
For VQA, we evaluate SOTA generalist models and our Shikra-13B onVQAv2 \citep{antol2015vqav2} and OK-VQA \citep{marino2019ok} following the normalization rules.
Here, we also provide VQAv2$^\text{val}$ (\textbf{83.3}) and OK-VQA (\textbf{53.8}) results on LVLM-eHub toolbox \citep{xu2023eHub} for easy comparison.
For Image Captioning, we evaluate them on COCO \citep{chen2015cococap} and Flickr30k \citep{plummer2015flickr30ke} in CIDEr.
We call Flamingo \citep{alayrac2022flamingo} FM for short.
}
\resizebox{\textwidth}{!}{%
\begin{tabular}{l|l|cccccccc}
\toprule
\multicolumn{2}{c|}{Datasets} & Shikra  & FM-80B & FM-9B & Kosmos-1 & BLIP-2 & Unified-IO  & VPGTrans & VisionLLM \\
\cmidrule(lr){1-2}\cmidrule(lr){3-10}
\multirow{4}{*}{VQA}
& VQAv2$^\text{val}$     & 75.33  &   -  &  -   & -    & 65.2 & -  & 65.2 & - \\
& VQAv2$^\text{dev}$     & 77.36  & 56.3 & 51.8 & 51.0 & 65.0 & 77.9  & - & - \\
& VQAv2$^\text{std}$     & 77.51  &  -   &  -   &  -   &  -  & -  & - & - \\
& OK-VQA                 & 47.16  & 50.6 & 44.7 &  -   & 45.9 & 54.0  & 45.0 & - \\
\cmidrule(lr){1-2}\cmidrule(lr){3-10}
\multirow{2}{*}{Caption}
& Flickr30k       & 73.9 &  67.2   &   61.5  & 67.1 &   -  &   -   &   -   & - \\
& COCO            & 117.5 & 84.3 & 79.4 & 84.7 &   -  & 122.3 &   -   & 114.2 \\
\bottomrule
\end{tabular}%
}
\label{tab:vl}
\end{table*}
\begin{table*}[h!]
\centering
\caption{\textbf{Object hallucination benchmark using POPE evaluation pipeline }\citep{li2023obj_Hallucination}.
Accuracy denotes the accuracy of predictions. Precision signifies the true positive samples among the predicted positives. Recall indicates the correct identification of all true positive samples. ``Yes'' represents the probability of the model outputting a positive answer.
Except for Shikra-7B, the other results are obtained from \citealp{li2023obj_Hallucination}.
}
\label{tab:pope_results}
\resizebox{\textwidth}{!}{%
\begin{tabular}{l|l|cccccc}
\toprule
Datasets & Metrics & Shikra & InstructBLIP  & MiniGPT-4 & LLaVA &MM-GPT & mPLUG-Owl \\
\cmidrule(lr){1-2}\cmidrule(lr){3-8}
\multirow{5}{*}{Random}
& Accuracy ($\uparrow$)       & 86.90 & 88.57 & 79.67 &50.37  & 50.10& 53.97 \\
& Precision ($\uparrow$)      & 94.40 & 84.09 &78.24  &50.19  & 50.05&52.07\\
& Recall ($\uparrow$)         & 79.27  & 95.13 &82.20  &  99.13& 100.00&99.60  \\
& F1-Score ($\uparrow$)       & 86.19 &89.27  &80.17  & 66.64 &  66.71 &68.39 \\
& Yes    & 43.26  & 56.57 & 52.53 & 98.77 &  99.90&95.63 \\
\cmidrule(lr){1-2}\cmidrule(lr){3-8}
\multirow{5}{*}{Popular}
& Accuracy ($\uparrow$)       & 83.97  & 82.77 &69.73  &49.87  & 50.00&50.90  \\
& Precision ($\uparrow$)      & 87.55 & 76.27 & 65.86 &49.93  & 50.00&50.46  \\
& Recall ($\uparrow$)         & 79.20  & 95.13 &81.93  & 99.27 & 100.00&99.40  \\
& F1-Score ($\uparrow$)       & 83.16 & 84.66 & 73.02 & 66.44 & 66.67 & 66.94\\
& Yes    & 45.23  & 62.37 & 62.20 & 99.40 &100.00  &98.57\\
\cmidrule(lr){1-2}\cmidrule(lr){3-8}
\multirow{5}{*}{Adversarial}
& Accuracy ($\uparrow$)       & 83.10  & 72.10  &65.17  &  49.70& 50.00 & 50.67\\
& Precision ($\uparrow$)      & 85.60  & 65.13 & 61.19 & 49.85 & 50.00 & 50.34\\
& Recall ($\uparrow$)         & 79.60 & 95.13 & 82.93 & 99.07 & 100.00 & 99.33\\
& F1-Score ($\uparrow$)       & 82.49 & 77.32 &  70.42& 66.32 &66.67  & 66.82\\
& Yes    & 46.50 & 73.03 &67.77  & 99.37 &   100.00&98.67\\
\bottomrule
\end{tabular}%
}
\end{table*}

For detect object in  autoregressive model,
several methods \citep{chen2021pix2seq,wang2022ofa} introduce extra vocabularies (\eg, <bin\_0>, $\cdots$, <bin\_1000>) to represent coordinates for object detection in spatially discretized images, as described in \secref{sec:pos_rep}.
In contrast, Shikra represents coordinates naturally and intuitively, using numbers directly.
Which form is better? We train two toy Shikra using two different representations with REC data, they performance is recorded in \tabref{tab:numer}, where using numbers directly achieves better results.  Aside from performance, our simple-designed coordinate numerical representation makes the model more elegant without modifying vocabularies for localization tasks. Users can freely control the precision of numerical representation (number of digits after the decimal separator) without retraining vocabularies. However, it also has drawbacks. Compared to using extra vocabularies, numerical representation requires more tokens to represent coordinates, leading to increased computational costs when predicting dense objects. In this paper, we still prefer numerical representation, but future research can choose the appropriate method based on their pros and cons.

\subsection{Quantitative results on conventional tasks}
\label{sec:experiment_vltask}
Our Shikra excels in Referential Dialogue, facilitating seamless integration into a wide range of vision-language (VL) tasks, particularly those related to positioning.
Here, we present the quantitative results for these tasks.

To demonstrate the positioning capability of our model,  we examine the REC task, in which models are ask to ground the object described with an expression.
As shown in \tabref{tab:rec}, we compare our method with generalist VL models that perform multiple tasks without finetuning. We also compare our method with Specialist SOTAs, including localization specialist models and generalist/foundation models that perform specific finetunes on localization-related tasks.
In this setting, we instruct Shikra to provide the coordinates of the objects referred to by the expression.
For an example, we use ``\textit{I'd like to know the exact coordinates of <expr> in the photo <image>.}'', where <expr> represents the expression and <image> represents the input image. More instructions can be found in \appref{tab:task_temp}.
The experimental results demonstrate that Shikra achieves promising performance compared to other generalist models.

Correspondingly, to quantitatively evaluate our model's understanding of position inputs, we evaluated our model on two types PointQA datasets, LookTwice-QA of \citep{mani2020pointqa} and Visual7W (PointQA Setting) of \citep{zhu2016v7w}.
LookTwice-QA asks models to answer questions about the region specified by the user, either by center point or box, with the distinction that these questions necessitate comprehending the user-designated area first, and then observing the entire image to answer.
For instance, ``\textit{How many of these} [Pronoun/Superclass/Class] \textit{<obj>?}'', where <obj> denotes the coordinates of input point or box and [Pronoun/Superclass/Class] represents language instructions with different clarity levels (\eg, [$\varnothing$/fruits/apples]).
Visual7W also provides a setting for point QA, where models are given a question and four box options, and should choose one as the answer.'
Our Shikra achieves the SOTA performance in all these settings.

Additionally, we assess our model on conventional VL tasks in \tabref{tab:vl}, such as VQA and Image Captioning, which do not necessitate coordinates in their input or output.
The experimental results show that we achieved promising results on most datasets. 
We also evaluated the performance of our method in POPE evalution pipeline \citep{li2023obj_Hallucination}, and the results are recorded in \tabref{tab:pope_results}. Our method has achieved results comparable to InstrutBLIP\citep{dai2023instructblip} and far surpasses recent popular MLLMs.
It's worth noting that these task configurations are just some subsets of Referential Dialogue. We hope readers can appreciate the more intriguing capabilities of Shikra in \figref{fig:cases} and \appref{sec:more_cases}.

\section{Limitations} 
\label{sec:limitations}
Shikra only supports English and is not user-friendly for non-English speakers. Making Shikra multilingual in the future is valuable. 
Shikra is unsuitable for dense object detection and segmentation tasks. Exploring improved coordinate representations for these tasks is also interesting. 
Shikra, like most LLMs, may produce harmful and counterfactual responses.


\section{Conclusion} 
\label{sec:conclusion}
Our study unveiled the critical gap in MLLMs' ability to understand and engage in referential dialogue, an integral aspect of human communication. To address this, we introduced Shikra, a unified, straightforward model designed to comprehend and output spatial coordinates in natural language. Our approach does not necessitate extra vocabularies, position encoders, or external plug-ins, preserving the model's simplicity. It was proved that Shikra performs notably well on a variety of conventional vision-language tasks, while offering swathes of exciting applications such as aiding AI assistants in Mixed Reality headsets or facilitating precise communication in online shopping scenery. 

\bibliography{anthology,custom}

\begin{thebibliography}{57}
\expandafter\ifx\csname natexlab\endcsname\relax\def\natexlab#1{#1}\fi

\bibitem[{Alayrac et~al.(2022)Alayrac, Donahue, Luc, Miech, Barr, Hasson, Lenc,
  Mensch, Millican, Reynolds et~al.}]{alayrac2022flamingo}
Jean-Baptiste Alayrac, Jeff Donahue, Pauline Luc, Antoine Miech, Iain Barr,
  Yana Hasson, Karel Lenc, Arthur Mensch, Katherine Millican, Malcolm Reynolds,
  et~al. 2022.
\newblock Flamingo: a visual language model for few-shot learning.
\newblock \emph{Advances in Neural Information Processing Systems},
  35:23716--23736.

\bibitem[{Antol et~al.(2015)Antol, Agrawal, Lu, Mitchell, Batra, Zitnick, and
  Parikh}]{antol2015vqav2}
Stanislaw Antol, Aishwarya Agrawal, Jiasen Lu, Margaret Mitchell, Dhruv Batra,
  C~Lawrence Zitnick, and Devi Parikh. 2015.
\newblock Vqa: Visual question answering.
\newblock In \emph{Proceedings of the IEEE international conference on computer
  vision}, pages 2425--2433.

\bibitem[{Awadalla et~al.(2023)Awadalla, Gao, Gardner, Hessel, Hanafy, Zhu,
  Marathe, Bitton, Gadre, Jitsev, Kornblith, Koh, Ilharco, Wortsman, and
  Schmidt}]{anas2023OpenFlamingo}
Anas Awadalla, Irena Gao, Joshua Gardner, Jack Hessel, Yusuf Hanafy, Wanrong
  Zhu, Kalyani Marathe, Yonatan Bitton, Samir Gadre, Jenia Jitsev, Simon
  Kornblith, Pang~Wei Koh, Gabriel Ilharco, Mitchell Wortsman, and Ludwig
  Schmidt. 2023.
\newblock \href {https://doi.org/10.5281/zenodo.7733589} {Openflamingo}.

\bibitem[{Bracha et~al.(2023)Bracha, Shaar, Shamsian, Fetaya, and
  Chechik}]{bracha2023disclip}
Lior Bracha, Eitan Shaar, Aviv Shamsian, Ethan Fetaya, and Gal Chechik. 2023.
\newblock Disclip: Open-vocabulary referring expression generation.
\newblock \emph{arXiv preprint arXiv:2305.19108}.

\bibitem[{Carion et~al.(2020)Carion, Massa, Synnaeve, Usunier, Kirillov, and
  Zagoruyko}]{carion2020detr}
Nicolas Carion, Francisco Massa, Gabriel Synnaeve, Nicolas Usunier, Alexander
  Kirillov, and Sergey Zagoruyko. 2020.
\newblock End-to-end object detection with transformers.
\newblock In \emph{Computer Vision--ECCV 2020: 16th European Conference,
  Glasgow, UK, August 23--28, 2020, Proceedings, Part I 16}, pages 213--229.
  Springer.

\bibitem[{Chen et~al.(2023)Chen, Han, Zhao, Zhang, Shi, Xu, and
  Xu}]{chen2023xllm}
Feilong Chen, Minglun Han, Haozhi Zhao, Qingyang Zhang, Jing Shi, Shuang Xu,
  and Bo~Xu. 2023.
\newblock {X-LLM}: Bootstrapping advanced large language models by treating
  multi-modalities as foreign languages.
\newblock \emph{arXiv preprint arXiv:2305.04160}.

\bibitem[{Chen et~al.(2021)Chen, Saxena, Li, Fleet, and
  Hinton}]{chen2021pix2seq}
Ting Chen, Saurabh Saxena, Lala Li, David~J Fleet, and Geoffrey Hinton. 2021.
\newblock Pix2seq: A language modeling framework for object detection.
\newblock \emph{arXiv preprint arXiv:2109.10852}.

\bibitem[{Chen et~al.(2015)Chen, Fang, Lin, Vedantam, Gupta, Doll{\'a}r, and
  Zitnick}]{chen2015cococap}
Xinlei Chen, Hao Fang, Tsung-Yi Lin, Ramakrishna Vedantam, Saurabh Gupta, Piotr
  Doll{\'a}r, and C~Lawrence Zitnick. 2015.
\newblock Microsoft coco captions: Data collection and evaluation server.
\newblock \emph{arXiv preprint arXiv:1504.00325}.

\bibitem[{Dai et~al.(2023)Dai, Li, Li, Tiong, Zhao, Wang, Li, Fung, and
  Hoi}]{dai2023instructblip}
Wenliang Dai, Junnan Li, Dongxu Li, Anthony Meng~Huat Tiong, Junqi Zhao,
  Weisheng Wang, Boyang Li, Pascale Fung, and Steven Hoi. 2023.
\newblock Instructblip: Towards general-purpose vision-language models with
  instruction tuning.
\newblock \emph{arXiv preprint arXiv:2305.06500}.

\bibitem[{Driess et~al.(2023)Driess, Xia, Sajjadi, Lynch, Chowdhery, Ichter,
  Wahid, Tompson, Vuong, Yu et~al.}]{driess2023palme}
Danny Driess, Fei Xia, Mehdi~SM Sajjadi, Corey Lynch, Aakanksha Chowdhery,
  Brian Ichter, Ayzaan Wahid, Jonathan Tompson, Quan Vuong, Tianhe Yu, et~al.
  2023.
\newblock {PaLM-E}: An embodied multimodal language model.
\newblock \emph{arXiv preprint arXiv:2303.03378}.

\bibitem[{Gao et~al.(2023)Gao, Han, Zhang, Lin, Geng, Zhou, Zhang, Lu, He, Yue
  et~al.}]{gao2023la_v2}
Peng Gao, Jiaming Han, Renrui Zhang, Ziyi Lin, Shijie Geng, Aojun Zhou, Wei
  Zhang, Pan Lu, Conghui He, Xiangyu Yue, et~al. 2023.
\newblock Llama-adapter v2: Parameter-efficient visual instruction model.
\newblock \emph{arXiv preprint arXiv:2304.15010}.

\bibitem[{Girshick(2015)}]{girshick2015fastrcnn}
Ross Girshick. 2015.
\newblock Fast r-cnn.
\newblock In \emph{Proceedings of the IEEE international conference on computer
  vision}, pages 1440--1448.

\bibitem[{Gong et~al.(2023)Gong, Lyu, Zhang, Wang, Zheng, Zhao, Liu, Zhang,
  Luo, and Chen}]{gong2023mmgpt}
Tao Gong, Chengqi Lyu, Shilong Zhang, Yudong Wang, Miao Zheng, Qian Zhao,
  Kuikun Liu, Wenwei Zhang, Ping Luo, and Kai Chen. 2023.
\newblock Multimodal-gpt: A vision and language model for dialogue with humans.
\newblock \emph{arXiv preprint arXiv:2305.04790}.

\bibitem[{Gupta et~al.(2019)Gupta, Dollar, and Girshick}]{gupta2019lvis}
Agrim Gupta, Piotr Dollar, and Ross Girshick. 2019.
\newblock Lvis: A dataset for large vocabulary instance segmentation.
\newblock In \emph{Proceedings of the IEEE/CVF conference on computer vision
  and pattern recognition}, pages 5356--5364.

\bibitem[{Hu et~al.(2017)Hu, Rohrbach, Andreas, Darrell, and
  Saenko}]{hu2017v7w_method1}
Ronghang Hu, Marcus Rohrbach, Jacob Andreas, Trevor Darrell, and Kate Saenko.
  2017.
\newblock Modeling relationships in referential expressions with compositional
  modular networks.
\newblock In \emph{CVPR}, pages 1115--1124.

\bibitem[{Huang et~al.(2023)Huang, Dong, Wang, Hao, Singhal, Ma, Lv, Cui,
  Mohammed, Liu et~al.}]{kosmos}
Shaohan Huang, Li~Dong, Wenhui Wang, Yaru Hao, Saksham Singhal, Shuming Ma,
  Tengchao Lv, Lei Cui, Owais~Khan Mohammed, Qiang Liu, et~al. 2023.
\newblock Language is not all you need: Aligning perception with language
  models.
\newblock \emph{arXiv preprint arXiv:2302.14045}.

\bibitem[{Johnson et~al.(2017)Johnson, Hariharan, Van Der~Maaten, Fei-Fei,
  Lawrence~Zitnick, and Girshick}]{johnson2017clevr}
Justin Johnson, Bharath Hariharan, Laurens Van Der~Maaten, Li~Fei-Fei,
  C~Lawrence~Zitnick, and Ross Girshick. 2017.
\newblock {CLEVR}: A diagnostic dataset for compositional language and
  elementary visual reasoning.
\newblock In \emph{CVPR}, pages 2901--2910.

\bibitem[{Kazemzadeh et~al.(2014)Kazemzadeh, Ordonez, Matten, and
  Berg}]{kazemzadeh2014refcoco}
Sahar Kazemzadeh, Vicente Ordonez, Mark Matten, and Tamara Berg. 2014.
\newblock Referitgame: Referring to objects in photographs of natural scenes.
\newblock In \emph{EMNLP}, pages 787--798.

\bibitem[{Kirillov et~al.(2023)Kirillov, Mintun, Ravi, Mao, Rolland, Gustafson,
  Xiao, Whitehead, Berg, Lo et~al.}]{kirillov2023sam}
Alexander Kirillov, Eric Mintun, Nikhila Ravi, Hanzi Mao, Chloe Rolland, Laura
  Gustafson, Tete Xiao, Spencer Whitehead, Alexander~C Berg, Wan-Yen Lo, et~al.
  2023.
\newblock Segment anything.
\newblock \emph{arXiv preprint arXiv:2304.02643}.

\bibitem[{Koh et~al.(2023)Koh, Salakhutdinov, and Fried}]{koh2023fromge}
Jing~Yu Koh, Ruslan Salakhutdinov, and Daniel Fried. 2023.
\newblock Grounding language models to images for multimodal generation.
\newblock \emph{arXiv preprint arXiv:2301.13823}.

\bibitem[{Krishna et~al.(2017)Krishna, Zhu, Groth, Johnson, Hata, Kravitz,
  Chen, Kalantidis, Li, Shamma et~al.}]{krishna2017visualgenome}
Ranjay Krishna, Yuke Zhu, Oliver Groth, Justin Johnson, Kenji Hata, Joshua
  Kravitz, Stephanie Chen, Yannis Kalantidis, Li-Jia Li, David~A Shamma, et~al.
  2017.
\newblock Visual {G}enome: Connecting language and vision using crowdsourced
  dense image annotations.
\newblock \emph{IJCV}, 123:32--73.

\bibitem[{Li et~al.(2023{\natexlab{a}})Li, Zhang, Chen, Wang, Yang, and
  Liu}]{li2023otter}
Bo~Li, Yuanhan Zhang, Liangyu Chen, Jinghao Wang, Jingkang Yang, and Ziwei Liu.
  2023{\natexlab{a}}.
\newblock Otter: A multi-modal model with in-context instruction tuning.
\newblock \emph{arXiv preprint arXiv:2305.03726}.

\bibitem[{Li et~al.(2023{\natexlab{b}})Li, Li, Savarese, and Hoi}]{li2023blip2}
Junnan Li, Dongxu Li, Silvio Savarese, and Steven Hoi. 2023{\natexlab{b}}.
\newblock Blip-2: Bootstrapping language-image pre-training with frozen image
  encoders and large language models.
\newblock \emph{arXiv preprint arXiv:2301.12597}.

\bibitem[{Li et~al.(2023{\natexlab{c}})Li, Du, Zhou, Wang, Zhao, and
  Wen}]{li2023obj_Hallucination}
Yifan Li, Yifan Du, Kun Zhou, Jinpeng Wang, Wayne~Xin Zhao, and Ji-Rong Wen.
  2023{\natexlab{c}}.
\newblock Evaluating object hallucination in large vision-language models.
\newblock \emph{arXiv preprint arXiv:2305.10355}.

\bibitem[{Lin et~al.(2020)Lin, Zhang, Chen, Cheng, and Lu}]{lin2020fca}
Zheng Lin, Zhao Zhang, Lin-Zhuo Chen, Ming-Ming Cheng, and Shao-Ping Lu. 2020.
\newblock Interactive image segmentation with first click attention.
\newblock In \emph{Proceedings of the IEEE/CVF conference on computer vision
  and pattern recognition}, pages 13339--13348.

\bibitem[{Lin et~al.(2022)Lin, Zhang, Han, and Lu}]{lin2022mmiis}
Zheng Lin, Zhao Zhang, Ling-Hao Han, and Shao-Ping Lu. 2022.
\newblock Multi-mode interactive image segmentation.
\newblock In \emph{Proceedings of the 30th ACM International Conference on
  Multimedia}, pages 905--914.

\bibitem[{Liu et~al.(2023{\natexlab{a}})Liu, Li, Wu, and Lee}]{liu2023llava}
Haotian Liu, Chunyuan Li, Qingyang Wu, and Yong~Jae Lee. 2023{\natexlab{a}}.
\newblock Visual instruction tuning.
\newblock \emph{arXiv preprint arXiv:2304.08485}.

\bibitem[{Liu et~al.(2017)Liu, Wang, and Yang}]{liu2017referring}
Jingyu Liu, Liang Wang, and Ming-Hsuan Yang. 2017.
\newblock Referring expression generation and comprehension via attributes.
\newblock In \emph{Proceedings of the IEEE International Conference on Computer
  Vision}, pages 4856--4864.

\bibitem[{Liu et~al.(2023{\natexlab{b}})Liu, Zeng, Ren, Li, Zhang, Yang, Li,
  Yang, Su, Zhu et~al.}]{liu2023gdino}
Shilong Liu, Zhaoyang Zeng, Tianhe Ren, Feng Li, Hao Zhang, Jie Yang, Chunyuan
  Li, Jianwei Yang, Hang Su, Jun Zhu, et~al. 2023{\natexlab{b}}.
\newblock Grounding dino: Marrying dino with grounded pre-training for open-set
  object detection.
\newblock \emph{arXiv preprint arXiv:2303.05499}.

\bibitem[{Loshchilov and Hutter(2017)}]{DBLP:conf/iclr/LoshchilovH17cos}
Ilya Loshchilov and Frank Hutter. 2017.
\newblock \href {https://openreview.net/forum?id=Skq89Scxx} {{SGDR:} stochastic
  gradient descent with warm restarts}.
\newblock In \emph{5th International Conference on Learning Representations,
  {ICLR} 2017, Toulon, France, April 24-26, 2017, Conference Track
  Proceedings}. OpenReview.net.

\bibitem[{Loshchilov and Hutter(2019)}]{DBLP:conf/iclr/LoshchilovH19adamw}
Ilya Loshchilov and Frank Hutter. 2019.
\newblock \href {https://openreview.net/forum?id=Bkg6RiCqY7} {Decoupled weight
  decay regularization}.
\newblock In \emph{7th International Conference on Learning Representations,
  {ICLR} 2019, New Orleans, LA, USA, May 6-9, 2019}. OpenReview.net.

\bibitem[{Lu et~al.(2022)Lu, Clark, Zellers, Mottaghi, and
  Kembhavi}]{lu2022unified}
Jiasen Lu, Christopher Clark, Rowan Zellers, Roozbeh Mottaghi, and Aniruddha
  Kembhavi. 2022.
\newblock Unified-io: A unified model for vision, language, and multi-modal
  tasks.
\newblock \emph{arXiv preprint arXiv:2206.08916}.

\bibitem[{Lu et~al.(2020)Lu, Goswami, Rohrbach, Parikh, and Lee}]{lu202012in1}
Jiasen Lu, Vedanuj Goswami, Marcus Rohrbach, Devi Parikh, and Stefan Lee. 2020.
\newblock 12-in-1: Multi-task vision and language representation learning.
\newblock In \emph{CVPR}, pages 10437--10446.

\bibitem[{Mani et~al.(2020)Mani, Yoo, Hinthorn, and
  Russakovsky}]{mani2020pointqa}
Arjun Mani, Nobline Yoo, Will Hinthorn, and Olga Russakovsky. 2020.
\newblock Point and ask: Incorporating pointing into visual question answering.
\newblock \emph{arXiv preprint arXiv:2011.13681}.

\bibitem[{Mao et~al.(2016)Mao, Huang, Toshev, Camburu, Yuille, and
  Murphy}]{mao2016refcocog}
Junhua Mao, Jonathan Huang, Alexander Toshev, Oana Camburu, Alan~L Yuille, and
  Kevin Murphy. 2016.
\newblock Generation and comprehension of unambiguous object descriptions.
\newblock In \emph{CVPR}, pages 11--20.

\bibitem[{Marino et~al.(2019)Marino, Rastegari, Farhadi, and
  Mottaghi}]{marino2019ok}
Kenneth Marino, Mohammad Rastegari, Ali Farhadi, and Roozbeh Mottaghi. 2019.
\newblock Ok-vqa: A visual question answering benchmark requiring external
  knowledge.
\newblock In \emph{Proceedings of the IEEE/cvf conference on computer vision
  and pattern recognition}, pages 3195--3204.

\bibitem[{Mu et~al.(2023)Mu, Zhang, Hu, Wang, Ding, Jin, Wang, Dai, Qiao, and
  Luo}]{mu2023embodiedgpt}
Yao Mu, Qinglong Zhang, Mengkang Hu, Wenhai Wang, Mingyu Ding, Jun Jin, Bin
  Wang, Jifeng Dai, Yu~Qiao, and Ping Luo. 2023.
\newblock Embodiedgpt: Vision-language pre-training via embodied chain of
  thought.
\newblock \emph{arXiv preprint arXiv:2305.15021}.

\bibitem[{OpenAI(2023)}]{openai2023gpt4}
OpenAI. 2023.
\newblock \href {http://arxiv.org/abs/2303.08774} {Gpt-4 technical report}.

\bibitem[{Plummer et~al.(2015)Plummer, Wang, Cervantes, Caicedo, Hockenmaier,
  and Lazebnik}]{plummer2015flickr30ke}
Bryan~A Plummer, Liwei Wang, Chris~M Cervantes, Juan~C Caicedo, Julia
  Hockenmaier, and Svetlana Lazebnik. 2015.
\newblock Flickr30k entities: Collecting region-to-phrase correspondences for
  richer image-to-sentence models.
\newblock In \emph{Proceedings of the IEEE international conference on computer
  vision}, pages 2641--2649.

\bibitem[{Su et~al.(2023)Su, Lan, Li, Xu, Wang, and Cai}]{su2023pandagpt}
Yixuan Su, Tian Lan, Huayang Li, Jialu Xu, Yan Wang, and Deng Cai. 2023.
\newblock Pandagpt: One model to instruction-follow them all.
\newblock \emph{arXiv preprint arXiv:2305.16355}.

\bibitem[{Tancik et~al.(2020)Tancik, Srinivasan, Mildenhall, Fridovich-Keil,
  Raghavan, Singhal, Ramamoorthi, Barron, and Ng}]{tancik2020fourier}
Matthew Tancik, Pratul Srinivasan, Ben Mildenhall, Sara Fridovich-Keil, Nithin
  Raghavan, Utkarsh Singhal, Ravi Ramamoorthi, Jonathan Barron, and Ren Ng.
  2020.
\newblock Fourier features let networks learn high frequency functions in low
  dimensional domains.
\newblock \emph{Advances in Neural Information Processing Systems},
  33:7537--7547.

\bibitem[{Tian et~al.(2019)Tian, Shen, Chen, and He}]{tian2019fcos}
Zhi Tian, Chunhua Shen, Hao Chen, and Tong He. 2019.
\newblock Fcos: Fully convolutional one-stage object detection.
\newblock In \emph{Proceedings of the IEEE/CVF international conference on
  computer vision}, pages 9627--9636.

\bibitem[{Wang et~al.(2021)Wang, Song, Li, Sun, Sun, and Zheng}]{wang2021end}
Jianfeng Wang, Lin Song, Zeming Li, Hongbin Sun, Jian Sun, and Nanning Zheng.
  2021.
\newblock End-to-end object detection with fully convolutional network.
\newblock In \emph{Proceedings of the IEEE/CVF conference on computer vision
  and pattern recognition}, pages 15849--15858.

\bibitem[{Wang et~al.(2022{\natexlab{a}})Wang, Yang, Hu, Li, Lin, Gan, Liu,
  Liu, and Wang}]{wang2022git}
Jianfeng Wang, Zhengyuan Yang, Xiaowei Hu, Linjie Li, Kevin Lin, Zhe Gan,
  Zicheng Liu, Ce~Liu, and Lijuan Wang. 2022{\natexlab{a}}.
\newblock Git: A generative image-to-text transformer for vision and language.
\newblock \emph{arXiv preprint arXiv:2205.14100}.

\bibitem[{Wang et~al.(2023{\natexlab{a}})Wang, Wang, Lin, Bai, Zhou, Zhou,
  Wang, and Zhou}]{wang2023one_peace}
Peng Wang, Shijie Wang, Junyang Lin, Shuai Bai, Xiaohuan Zhou, Jingren Zhou,
  Xinggang Wang, and Chang Zhou. 2023{\natexlab{a}}.
\newblock {ONE-PEACE}: Exploring one general representation model toward
  unlimited modalities.
\newblock \emph{arXiv preprint arXiv:2305.11172}.

\bibitem[{Wang et~al.(2022{\natexlab{b}})Wang, Yang, Men, Lin, Bai, Li, Ma,
  Zhou, Zhou, and Yang}]{wang2022ofa}
Peng Wang, An~Yang, Rui Men, Junyang Lin, Shuai Bai, Zhikang Li, Jianxin Ma,
  Chang Zhou, Jingren Zhou, and Hongxia Yang. 2022{\natexlab{b}}.
\newblock Ofa: Unifying architectures, tasks, and modalities through a simple
  sequence-to-sequence learning framework.
\newblock In \emph{International Conference on Machine Learning}, pages
  23318--23340. PMLR.

\bibitem[{Wang et~al.(2023{\natexlab{b}})Wang, Chen, Chen, Wu, Zhu, Zeng, Luo,
  Lu, Zhou, Qiao et~al.}]{wang2023visionllm}
Wenhai Wang, Zhe Chen, Xiaokang Chen, Jiannan Wu, Xizhou Zhu, Gang Zeng, Ping
  Luo, Tong Lu, Jie Zhou, Yu~Qiao, et~al. 2023{\natexlab{b}}.
\newblock Visionllm: Large language model is also an open-ended decoder for
  vision-centric tasks.
\newblock \emph{arXiv preprint arXiv:2305.11175}.

\bibitem[{Xie et~al.(2023)Xie, Zhang, Wu, Zhu, Zhao, and Liang}]{xie2023dod}
Chi Xie, Zhao Zhang, Yixuan Wu, Feng Zhu, Rui Zhao, and Shuang Liang. 2023.
\newblock Exposing the troublemakers in described object detection.
\newblock \emph{arXiv preprint}.

\bibitem[{Xu et~al.(2023)Xu, Shao, Zhang, Gao, Liu, Lei, Meng, Huang, Qiao, and
  Luo}]{xu2023eHub}
Peng Xu, Wenqi Shao, Kaipeng Zhang, Peng Gao, Shuo Liu, Meng Lei, Fanqing Meng,
  Siyuan Huang, Yu~Qiao, and Ping Luo. 2023.
\newblock Lvlm-ehub: A comprehensive evaluation benchmark for large
  vision-language models.
\newblock \emph{arXiv preprint arXiv:2306.09265}.

\bibitem[{Yan et~al.(2023)Yan, Jiang, Wu, Wang, Luo, Yuan, and
  Lu}]{yan2023uninext}
Bin Yan, Yi~Jiang, Jiannan Wu, Dong Wang, Ping Luo, Zehuan Yuan, and Huchuan
  Lu. 2023.
\newblock Universal instance perception as object discovery and retrieval.
\newblock In \emph{Proceedings of the IEEE/CVF Conference on Computer Vision
  and Pattern Recognition}, pages 15325--15336.

\bibitem[{Yang et~al.(2022)Yang, Gan, Wang, Hu, Ahmed, Liu, Lu, and
  Wang}]{yang2022unitab}
Zhengyuan Yang, Zhe Gan, Jianfeng Wang, Xiaowei Hu, Faisal Ahmed, Zicheng Liu,
  Yumao Lu, and Lijuan Wang. 2022.
\newblock Unitab: Unifying text and box outputs for grounded vision-language
  modeling.
\newblock In \emph{Computer Vision--ECCV 2022: 17th European Conference, Tel
  Aviv, Israel, October 23--27, 2022, Proceedings, Part XXXVI}, pages 521--539.
  Springer.

\bibitem[{Ye et~al.(2023)Ye, Xu, Xu, Ye, Yan, Zhou, Wang, Hu, Shi, Shi
  et~al.}]{ye2023mplug}
Qinghao Ye, Haiyang Xu, Guohai Xu, Jiabo Ye, Ming Yan, Yiyang Zhou, Junyang
  Wang, Anwen Hu, Pengcheng Shi, Yaya Shi, et~al. 2023.
\newblock mplug-owl: Modularization empowers large language models with
  multimodality.
\newblock \emph{arXiv preprint arXiv:2304.14178}.

\bibitem[{Zhang et~al.(2023{\natexlab{a}})Zhang, Fei, Yao, Ji, Li, Liu, and
  Chua}]{zhang2023transfer}
Ao~Zhang, Hao Fei, Yuan Yao, Wei Ji, Li~Li, Zhiyuan Liu, and Tat-Seng Chua.
  2023{\natexlab{a}}.
\newblock Transfer visual prompt generator across llms.
\newblock \emph{arXiv preprint arXiv:2305.01278}.

\bibitem[{Zhang et~al.(2023{\natexlab{b}})Zhang, Zhang, Li, Zhao, Karypis, and
  Smola}]{zhang2023mmcot}
Zhuosheng Zhang, Aston Zhang, Mu~Li, Hai Zhao, George Karypis, and Alex Smola.
  2023{\natexlab{b}}.
\newblock Multimodal chain-of-thought reasoning in language models.
\newblock \emph{arXiv preprint arXiv:2302.00923}.

\bibitem[{Zhou et~al.(2020)Zhou, Wang, Liu, Hu, and Zhang}]{zhou2020more}
Yuanen Zhou, Meng Wang, Daqing Liu, Zhenzhen Hu, and Hanwang Zhang. 2020.
\newblock More grounded image captioning by distilling image-text matching
  model.
\newblock In \emph{Proceedings of the IEEE/CVF conference on computer vision
  and pattern recognition}, pages 4777--4786.

\bibitem[{Zhu et~al.(2023)Zhu, Chen, Shen, Li, and Elhoseiny}]{zhu2023minigpt}
Deyao Zhu, Jun Chen, Xiaoqian Shen, Xiang Li, and Mohamed Elhoseiny. 2023.
\newblock Minigpt-4: Enhancing vision-language understanding with advanced
  large language models.
\newblock \emph{arXiv preprint arXiv:2304.10592}.

\bibitem[{Zhu et~al.(2016)Zhu, Groth, Bernstein, and Fei-Fei}]{zhu2016v7w}
Yuke Zhu, Oliver Groth, Michael Bernstein, and Li~Fei-Fei. 2016.
\newblock Visual7w: Grounded question answering in images.
\newblock In \emph{Proceedings of the IEEE conference on computer vision and
  pattern recognition}, pages 4995--5004.

\end{thebibliography}
\bibliographystyle{acl_natbib}

\appendix
\label{sec:appendix}

\section{Details of All Training Data}
\label{sec:train_data}
\begin{table*}[t]
\centering
\renewcommand{\tabcolsep}{10mm}
\caption{\textbf{All training data used by Shikra.} The asterisk indicates that this data is only used in the second stage.}
\begin{tabular}{l|l}
\toprule
\textbf{Task} & \textbf{Dataset}  \\ \cmidrule(lr){1-2}
Captioning       &  LLaVA-Pretraining     \\ \cmidrule(lr){1-2}
Soptting Cap.   &  Flickr30K Entities \\ \cmidrule(lr){1-2}
Grounding Cap.   &  Visual Genome \\ \cmidrule(lr){1-2} 
REG   &  RefCOCO, RefCOCO+, RefCOCOg \\ \cmidrule(lr){1-2}
REC   &  RefCOCO, RefCOCO+, RefCOCOg, Visual Genome \\ \cmidrule(lr){1-2} 
VQA         &  VQAv2    \\ \cmidrule(lr){1-2}
PointQA     &  PointQA-Local/Twice, Visual-7W (`which box' subset)      \\ \cmidrule(lr){1-2}
Dialogue  & LLaVA-Instruct-150K*  \\ \cmidrule(lr){1-2} 
RD   & VCR, Shikra-RD (Generated data from Flickr30K Entities)* \\ 

\bottomrule
\end{tabular}%
\label{tab:all_data}
\end{table*}
We listed all training data in \tabref{tab:all_data}. 
The asterisk indicates that this data is only used in the second training stage.
We removed the images from the training set that are the same as those in the testing or validation set to prevent potential data leakage.

\section{Examples of Task Prompts}
\label{sec:task_temp}
\begin{table*}[h!]
\centering
\caption{
\textbf{Examples of task templates used by Shikra on different types of training data.}
The explanation of placeholders in the template is as follows:`<image>' represents the input image; `<objs>' refers to the center points or bounding box of a user-specified location; `<question>' denotes the question in the VQA dataset;  `<expr>' represents the expression in the REC task.
During inference, there is no need to be confined to these forms. Users can describe their needs in natural language, creating more diverse and engaging task formats.
}
\resizebox{\textwidth}{!}{%
\begin{tabular}{l|l}
\toprule
\textbf{Task} & 
\textbf{Three randomly chosen examples from hundreds.}  \\
\cmidrule(lr){1-1}\cmidrule(lr){2-2}
\multirow{3}{*}{Captioning}
& Describe this image <image> as simply as possible.  \\
& What is the content of the image <image>? Please answer in short sentences.    \\
& Summarize the content of the photo <image>.  \\
\cmidrule(lr){1-1}\cmidrule(lr){2-2}
\multirow{3}{*}{Spotting Cap.}
& Can you provide a description of the image <image> and include the coordinates [x0,y0,x1,y1] for each mentioned object?  \\
& Please explain what's happening in the photo <image> and give coordinates [xmin,ymin,xmax,ymax] for the items you reference.    \\
& How would you describe the contents of the image <image>? Please provide the positions of mentioned objects in square brackets.  \\
\cmidrule(lr){1-1}\cmidrule(lr){2-2}
\multirow{3}{*}{Grounding Cap.}
& Can you give me a description of the region <objs> in image <image>?  \\
& Describe what's happening within the coordinates <objs> of the given image <image>.    \\
& What does the area <objs> within the given visual <image> contain?  \\
\cmidrule(lr){1-1}\cmidrule(lr){2-2}
\multirow{3}{*}{REG}
& For the given image <image>, can you provide a unique description of the area <objs>?  \\
& In the photo <image>, how would you describe the selected area <objs> uniquely?    \\
& Can you provide a description for the region <objs> in the image <image> such that it sets it apart from others?  \\
\cmidrule(lr){1-1}\cmidrule(lr){2-2}
\multirow{3}{*}{$\text{Q} \rightarrow \text{A}$}
& I want to know the answer to `<question>' Refer to the image <image> and give a clear response.  \\
& Answer this question directly after referring to the image <image>: <question>    \\
& Examine the image <image> and provide a brief answer for `<question>'  \\
\cmidrule(lr){1-1}\cmidrule(lr){2-2}
\multirow{3}{*}{Q$\rightarrow$CA}
& Having a look at image <image>, can you tell me the answer to my question '<question>' and the logic leading to it?  \\
& Please answer the following question '<question>' based on the image <image>, and describe your thought process \\
& Upon analyzing the image <image>, please find the answer to my question '<question>' and provide a detailed explanation. \\
\cmidrule(lr){1-1}\cmidrule(lr){2-2}
\multirow{3}{*}{Q$\rightarrow$$\text{C}^\text{Point}$A}
& Analyze the image <image> and answer `<question>' Include your reasoning process and mark center points of related objects as [cx, cy].  \\
& Based on <image>, please respond to `<question>' Include your thought process and note involved objects using [cx, cy] for their center points.  \\
& While observing image <image>, kindly answer `<question>' Elaborate on your reasoning process and tag any object center points involved [x,y].\\
\cmidrule(lr){1-1}\cmidrule(lr){2-2}
\multirow{3}{*}{Q$\rightarrow$$\text{C}^\text{Box}$A}
& <question> Please offer your reasoning process, and provide bounding boxes of mentioned objects within square brackets. Here is the picture <image>  \\
& Please explain your reasoning and provide bounding boxes, denoted by square brackets, for the objects mentioned in the picture <image>. <question>  \\
& Consider the image <image>, and then provide a well-reasoned answer to the question '<question>' Don't forget to mark relevant object locations using [x0,y0,x1,y1].\\
\cmidrule(lr){1-1}\cmidrule(lr){2-2}
\multirow{3}{*}{REC}
& In the given <image>, could you find and tell me the coordinates of <expr>?  \\
& I need the coordinates of <expr> in <image>, can you please assist me with that?  \\
& Locate <expr> in <image> and provide its coordinates, please.\\
\bottomrule
\end{tabular}%
}
\label{tab:task_temp}
\end{table*}
We list some task prompts used by Shikra during training in \tabref{tab:task_temp}.
For every task listed, there are hundreds. These prompts are generated by GPT-4 with carefully designed instructions. We randomly selected three prompts for readers' better understanding.
Note that during inference, there is no need to confine oneself to these forms. Users can express their needs in natural language, creating diverse and engaging task formats.


\section{More Conversations with Shikra}
\label{sec:more_cases}
\begin{figure*}[t]
    \begin{center}
        \includegraphics[width=0.97\textwidth]{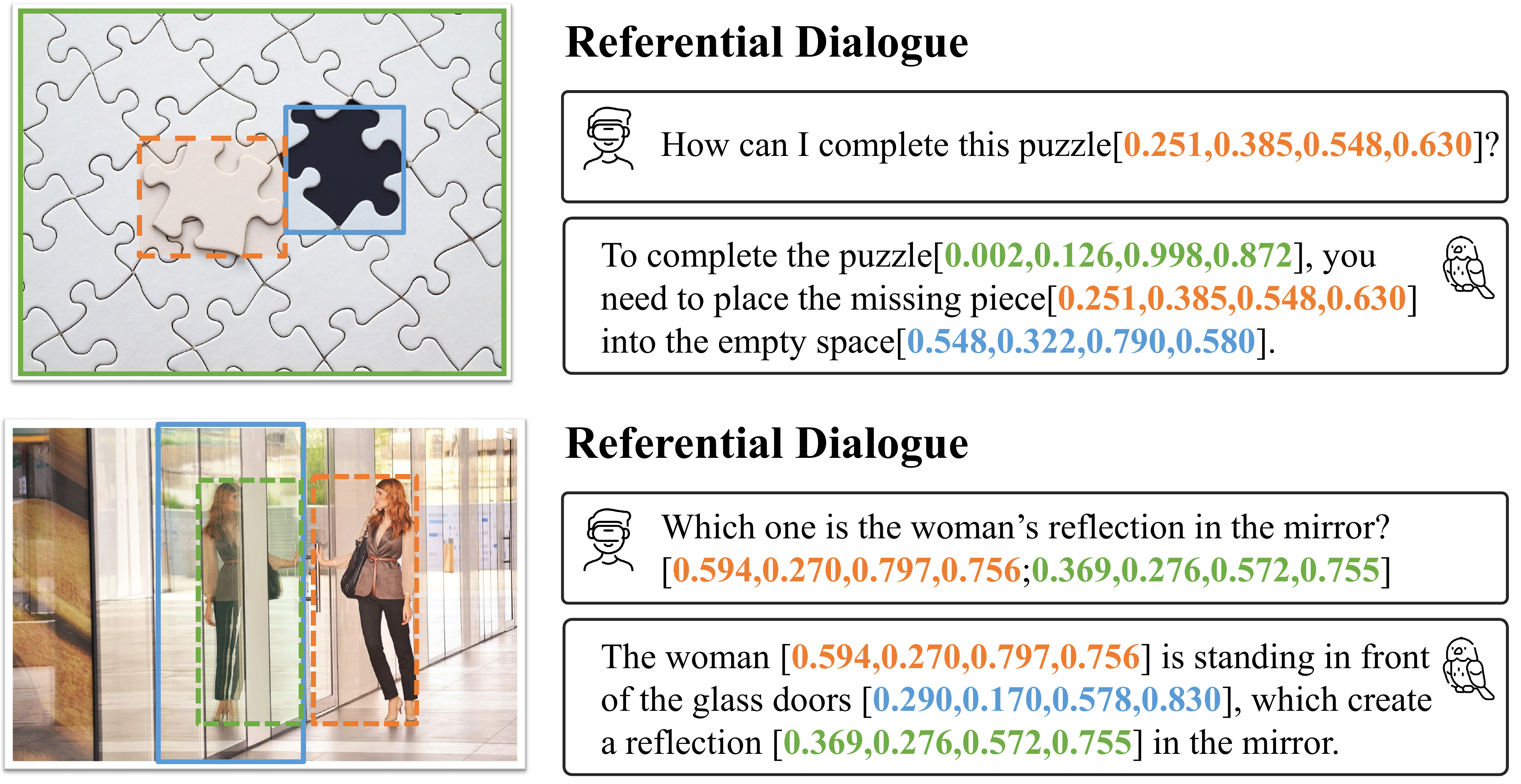}
    \end{center}
   \caption{\textbf{Referential Dialogue using Shikra-7B}.The dashed box on an image represents the area referred to by the user or jointly referred to by Shikra, while the solid box represents the area solely referred to by Shikra.}
\label{fig:case_rd}
\end{figure*}

\begin{figure*}[t]
    \begin{center}
        \includegraphics[width=0.7\textwidth]{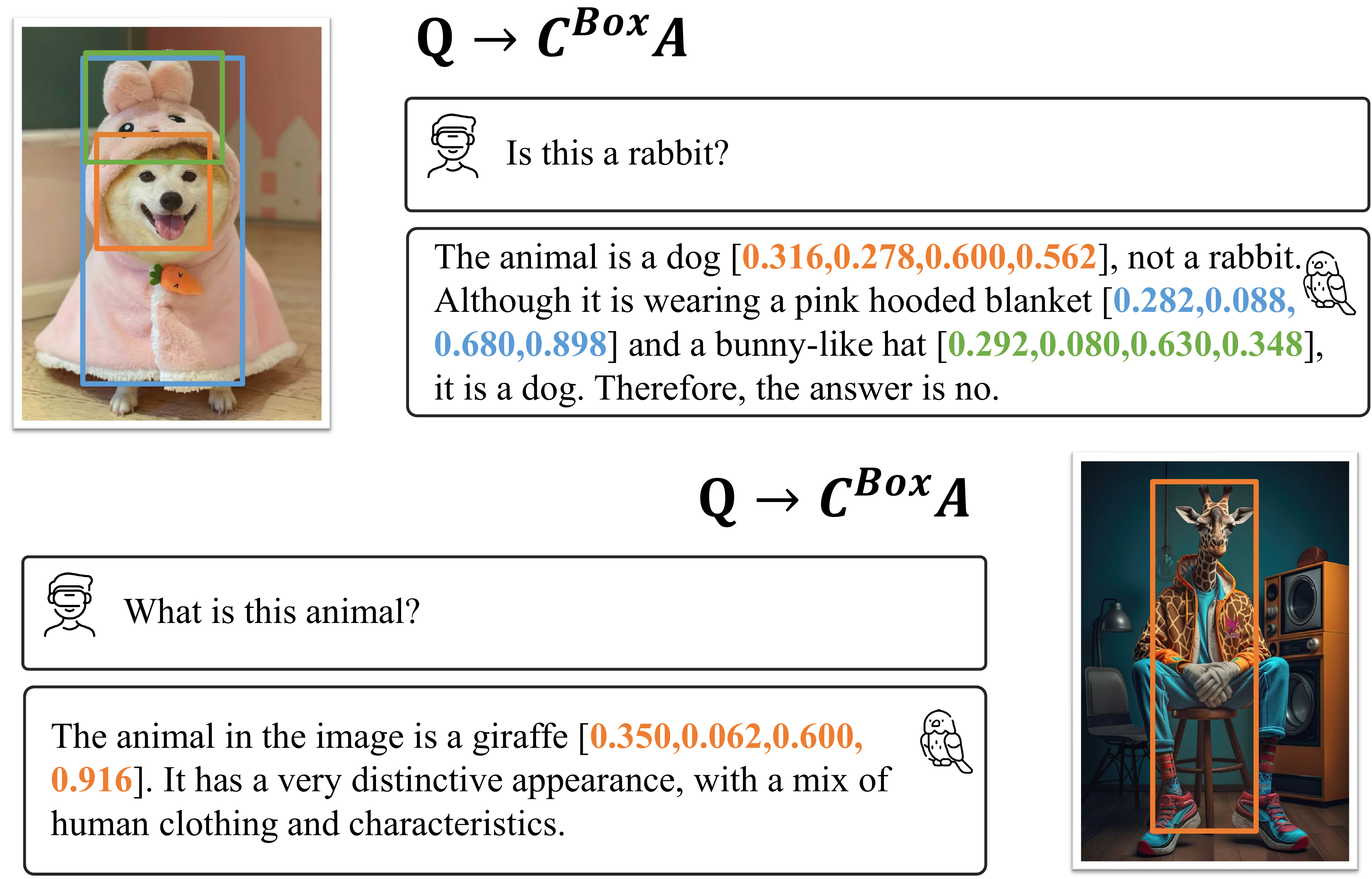}
    \end{center}
   \caption{\textbf{Q$\rightarrow$$\text{C}^\text{Box}$A using Shikra-7B}. It asks models generate grounded explanation for the answer.}
\label{fig:case_vqa}
\end{figure*}

\begin{figure*}[t]
    \begin{center}
        \includegraphics[width=0.6\textwidth]{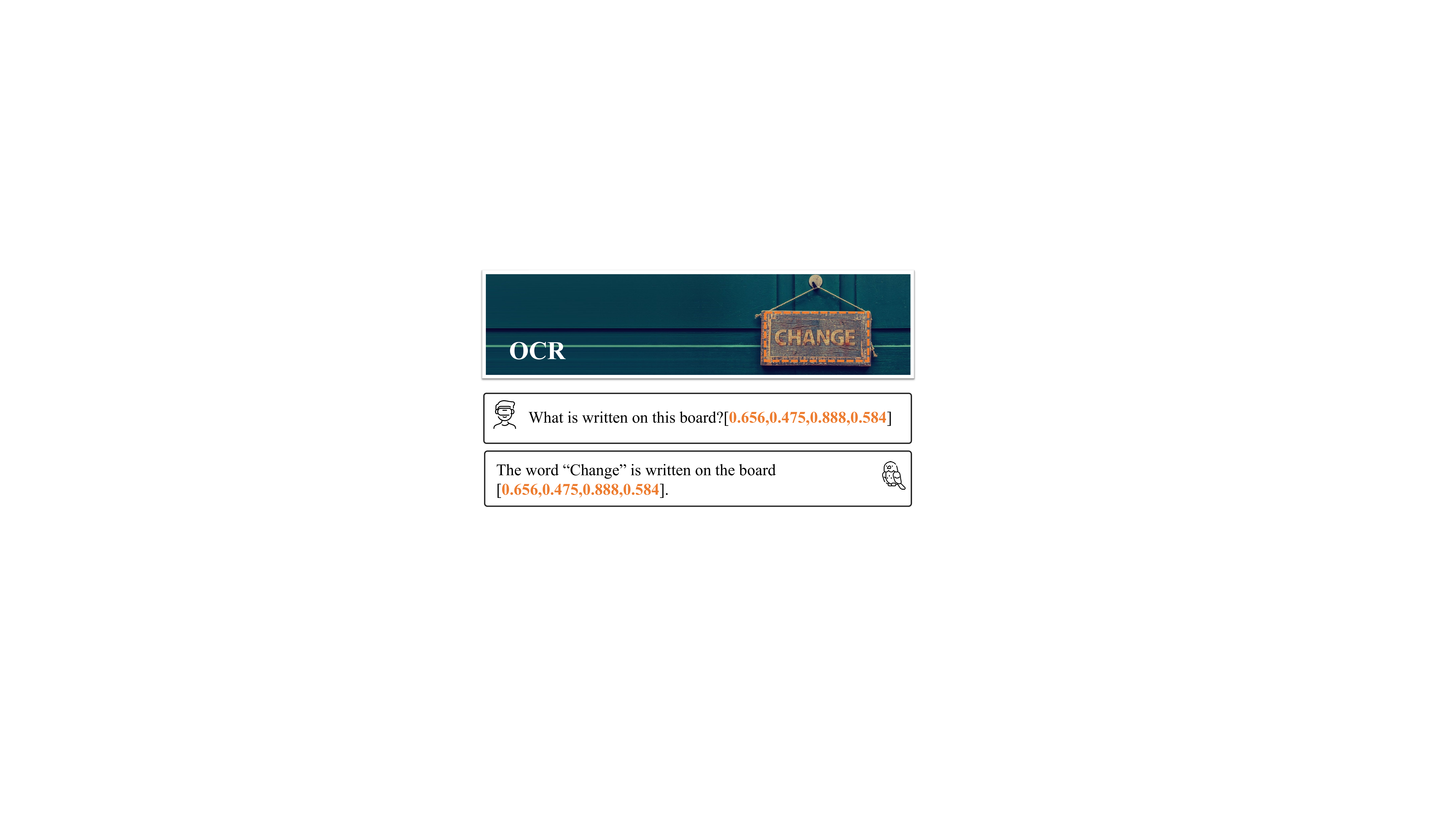}
    \end{center}
   \caption{\textbf{OCR using Shikra-7B}. We do not have explicit OCR datasets in Shikra training.}
\label{fig:case_ocr}
\end{figure*}

\begin{figure*}[t]
    \begin{center}
        \includegraphics[width=0.96\textwidth]{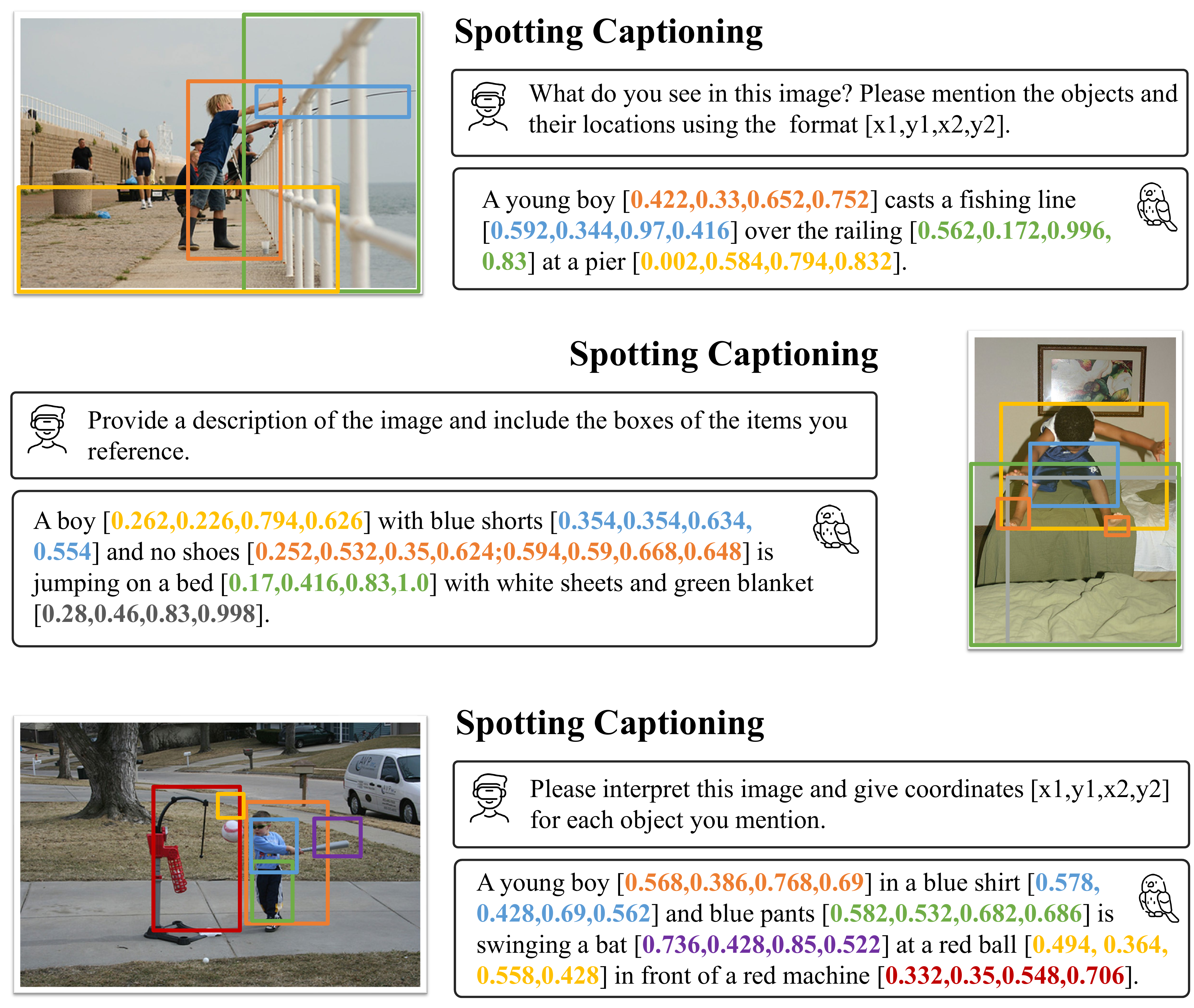}
    \end{center}
   \caption{\textbf{Spotting Captioning using Shikra-7B}. The task requires the model to describe the image and spots the mentioned objects or regions using points or boxes.}
\label{fig:case_spotting_captioning}
\end{figure*}

\begin{figure*}[t]
    \begin{center}
        \includegraphics[width=1.0\textwidth]{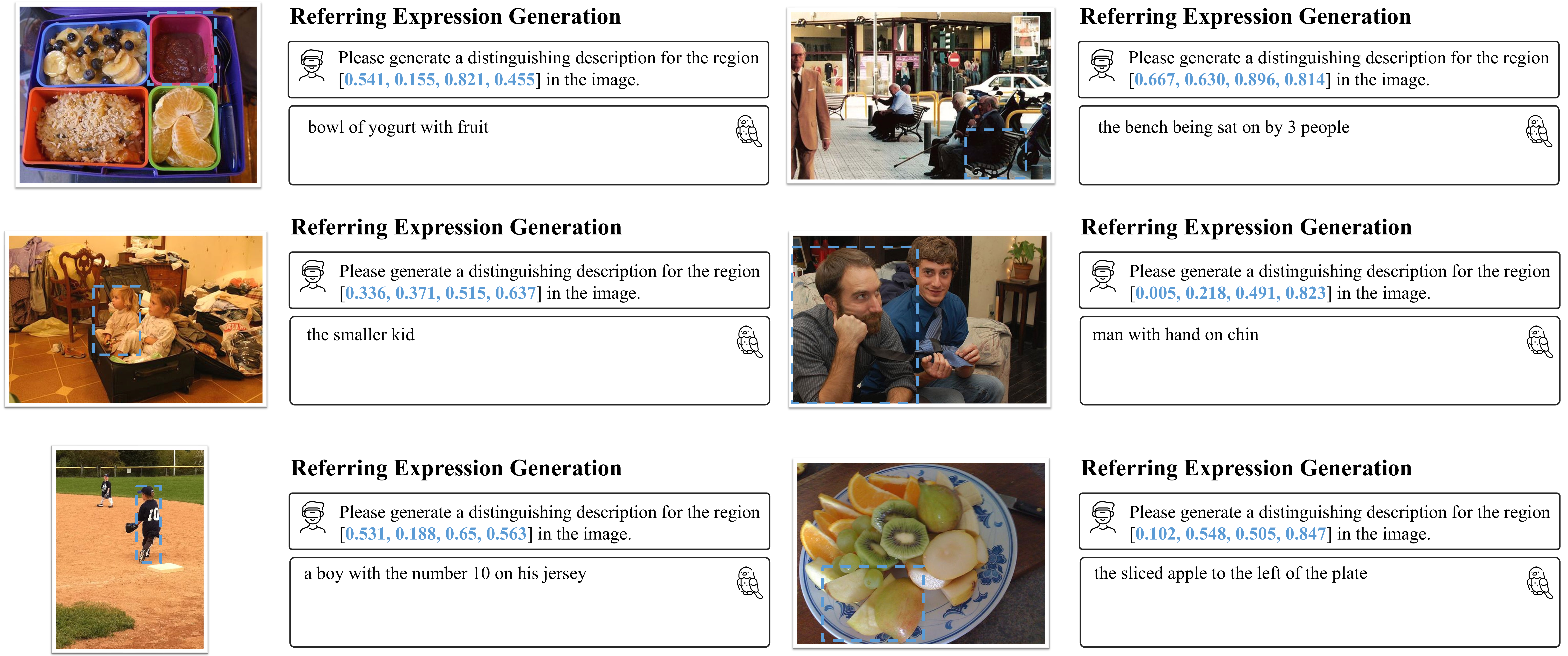}
    \end{center}
   \caption{\textbf{Referring Expression Generation (REG) using Shikra-7B}. The purpose of REG is to generate a unique description for a specified location.}
\label{fig:case_reg}
\end{figure*}

\begin{figure*}[t]
    \begin{center}
        \includegraphics[width=1.0\textwidth]{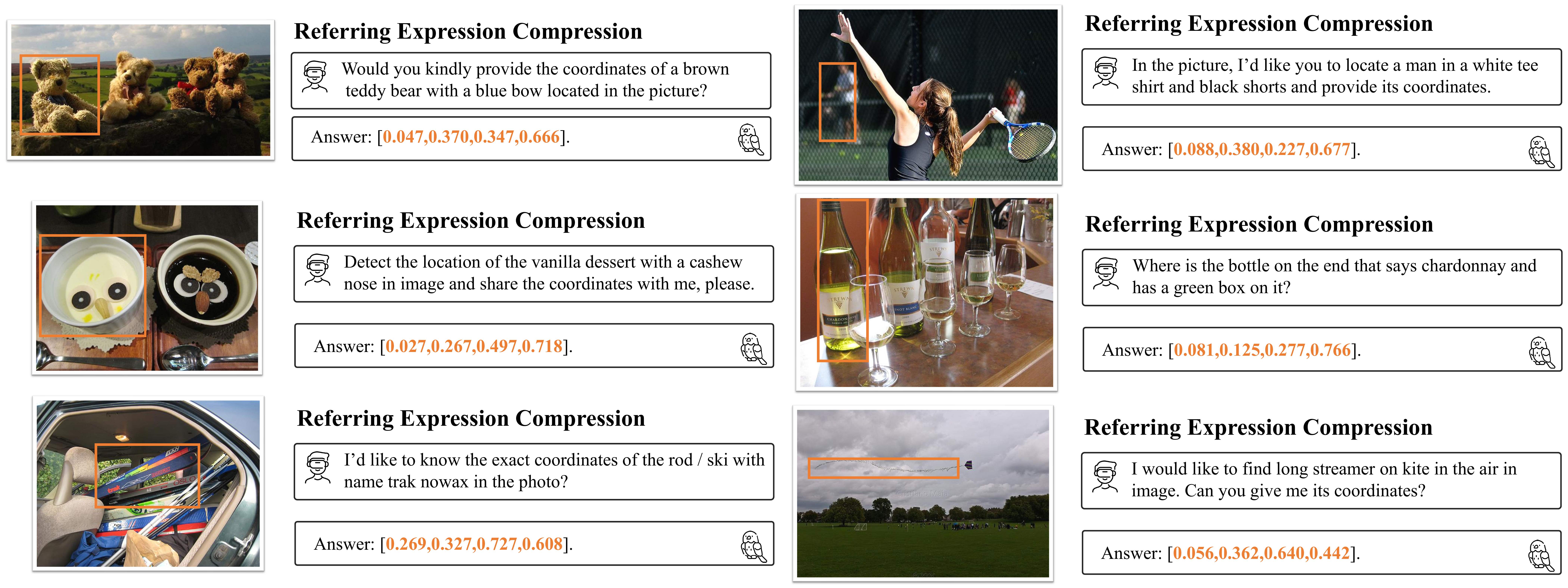}
    \end{center}
   \caption{\textbf{Referring Expression Comprehension (REC) using Shikra-7B}. The task aims to localize a target object in an image described by a referring expression.}
\label{fig:case_rec}
\end{figure*}

\begin{figure*}[t]
    \begin{center}
        \includegraphics[width=0.97\textwidth]{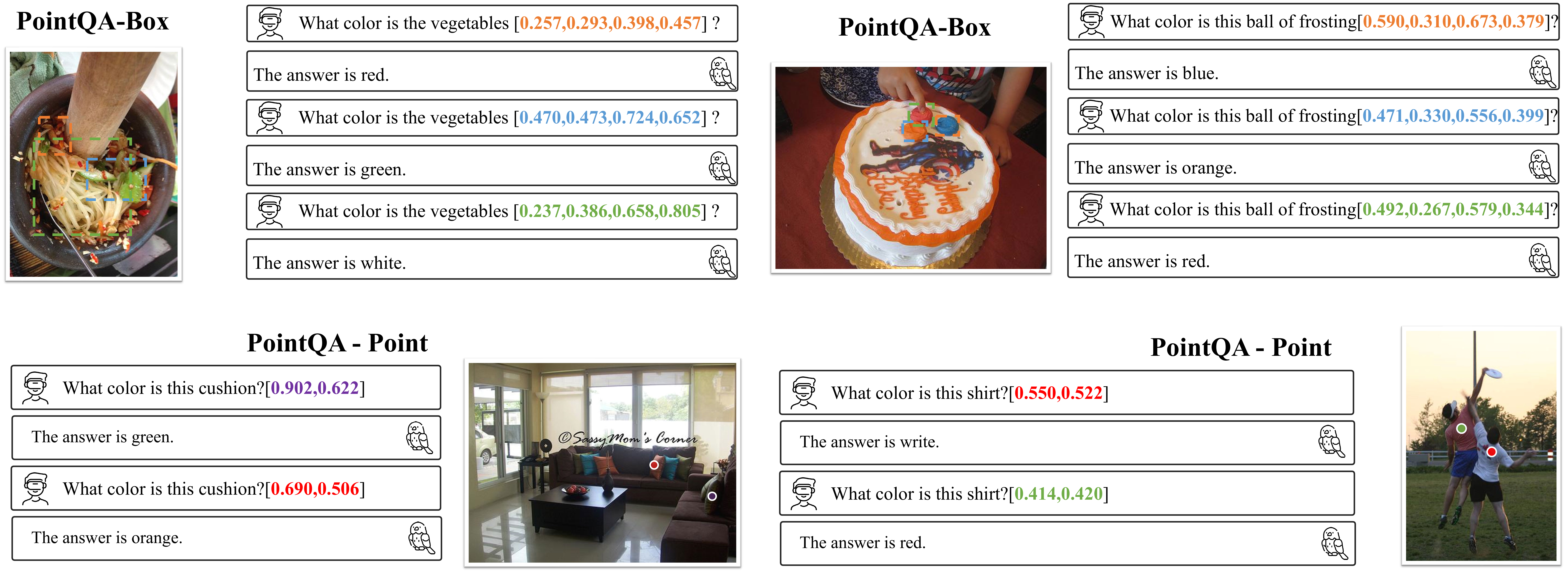}
    \end{center}
   \caption{\textbf{PointQA using Shikra-7B}. The task asks models to answer questions about the region specified by the user, either by center point or box.}
\label{fig:case_pointqa}
\end{figure*}

\begin{figure*}[t]
    \begin{center}
        \includegraphics[width=0.97\textwidth]{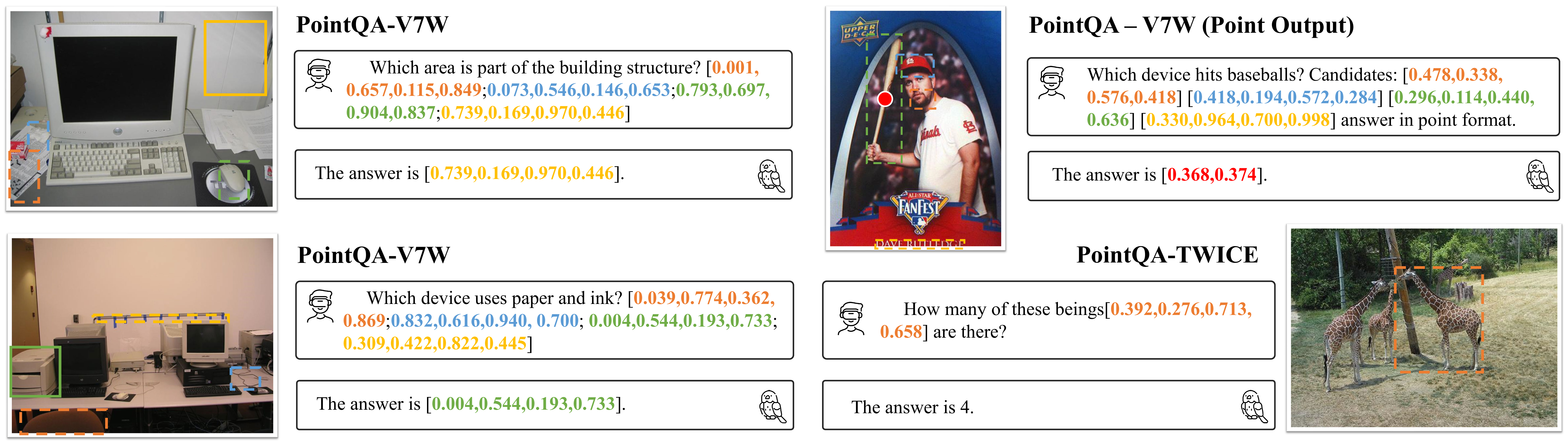}
    \end{center}
   \caption{\textbf{PointQA-V7W using Shikra-7B}.
PointQA-V7W provides a setting for point QA, where models are
given a question and four box options, and should
choose one as the answer.}
\label{fig:case_v7w}
\end{figure*}

We provide additional dialogue records of Shikra-7B in this section. For instance, we showcase RD results in \figref{fig:case_rd}, VQA (Q$\rightarrow$$\text{C}^\text{Box}$A) in \figref{fig:case_vqa}, and Spotting Captioning in \figref{fig:case_spotting_captioning}.
We also include examples of traditional VL task forms, like OCR in \figref{fig:case_ocr}, REC in \figref{fig:case_rec}, REG in \figref{fig:case_reg}, and PointQA in \figref{fig:case_pointqa}.
Furthermore, \figref{fig:case_pointqa} and \figref{fig:case_v7w} demonstrates that our input and output can handle points and boxes, just tell Shikra what to do.

\end{document}